\documentclass[lettersize,journal,twoside]{IEEEtran}
\usepackage{amsmath,amsfonts,amsthm,bm} % 
\usepackage{algorithmic}
\usepackage{array}
\usepackage[caption=false,font=normalsize,labelfont=sf,textfont=sf]{subfig}
\usepackage{textcomp}
\usepackage{url}
\usepackage{verbatim}
\usepackage{graphicx}
\def\BibTeX{{\rm B\kern-.05em{\sc i\kern-.025em b}\kern-.08em
    T\kern-.1667em\lower.7ex\hbox{E}\kern-.125emX}}
\usepackage{balance}

\makeatletter
\newcommand\dlmu[2][4cm]{\hskip1pt\underline{\hb@xt@ #1{\hss#2\hss}}\hskip3pt}
\makeatother

\usepackage{colortbl}
\usepackage{graphics}
\usepackage{epsfig}
\usepackage[ruled]{algorithm2e}
\usepackage{times}
\usepackage{amsmath}
\usepackage{amssymb}
\usepackage{multirow}
\usepackage{color}
\usepackage{dblfloatfix}
\usepackage{autobreak}
\usepackage{makecell} 
\usepackage{hyperref}
\usepackage[dvipsnames, svgnames, x11names,table]{xcolor}

 \usepackage{threeparttable}

\newcommand{\myfootnote}[1]{%
  \begingroup%
  \unrestored@protected@xdef\@thefnmark{\thempfn}%
  \footnotetext{#1}%
  \endgroup%
}

\newcommand{\revise}[1]{\textcolor{black}{#1}}
\newcommand\revisesec[1]{\textcolor{black}{#1}}
\makeatletter
\newcommand{\removelatexerror}{\let\@latex@error\@gobble}
\makeatother

\begin{document}
\title{\LARGE \bf
	BEVPlace++: Fast, Robust, and Lightweight LiDAR Global Localization for Unmanned Ground Vehicles
}

\author{Lun Luo$^{1,2}$, Si-Yuan Cao$^{3}$, Xiaorui Li$^{4}$, Jintao Xu$^{2}$, Rui Ai$^{2}$, Zhu Yu,$^{5}$ Xieyuanli Chen$^{1*}$
    \thanks{Manuscript received: January 2, 2025; Revised: May 6, 2025; Accepted: June 23, 2025.}
    \thanks{This paper was recommended for publication by Editor Javier Civera upon evaluation of the Associate Editor and Reviewers’ comments.}
    \thanks{$^{1}$L. Luo and X. Chen are with the College of Intelligence Science and Technology, National University of Defense Technology. $^{2}$L. Luo, J. Xu, and R. Ai are with Haomo.AI Technology Co., Ltd.
    $^{3}$S. Cao is with Ningbo Innovation Center, Zhejiang University. $^{4}$X. Li is with the College of Instrument Science and Optoelectronics Engineering, Beihang University. $^{5}$Z. Yu is with the College of Information Science and Electronic Engineering, Zhejiang University.}
    \thanks{*Corresponding author: Xieyuanli Chen (xieyuanli.chen@nudt.edu.cn).}
    \thanks{This work was supported in part by the National Science Foundation of China under Grant 62403478 and the Young Elite Scientists Sponsorship Program by CAST (No. 2023QNRC001).}
    \thanks{Digital Object Identifier (DOI): see top of this page.}
}

\markboth{IEEE TRANSACTIONS ON ROBOTICS. PREPRINT VERSION. ACCEPTED JUNE 2025}%
{Luo \MakeLowercase{\textit{et al.}}: BEVPlace++: Fast, Robust, and Lightweight LiDAR Global Localization for Unmanned Ground Vehicles}

\maketitle
\begin{abstract}
This article introduces BEVPlace++, a novel, fast, and robust LiDAR global localization method for unmanned ground vehicles. It uses lightweight convolutional neural networks (CNNs) on Bird's Eye View (BEV) image-like representations of LiDAR data to achieve accurate global localization through place recognition, followed by 3-DoF pose estimation. Our detailed analyses reveal an interesting fact that CNNs are inherently effective at extracting distinctive features from LiDAR BEV images. Remarkably, keypoints of two BEV images with large translations can be effectively matched using CNN-extracted features. Building on this insight, we design a Rotation Equivariant Module (REM) to obtain distinctive features while enhancing robustness to rotational changes. A Rotation Equivariant and Invariant Network (REIN) is then developed by cascading REM and a descriptor generator, NetVLAD, to sequentially generate rotation equivariant local features and rotation invariant global descriptors. The global descriptors are used first to achieve robust place recognition, and then local features are used for accurate pose estimation. \revise{Experimental results on seven public datasets and our UGV platform demonstrate that BEVPlace++, even when trained on a small dataset (3000 frames of KITTI) only with place labels, generalizes well to unseen environments, performs consistently across different days and years, and adapts to various types of LiDAR scanners.} BEVPlace++ achieves state-of-the-art performance in multiple tasks, including place recognition, loop closure detection, and global localization. 
Additionally, BEVPlace++ is lightweight, runs in real-time, and does not require accurate pose supervision, making it highly convenient for deployment. \revise{The source codes are publicly available at \href{https://github.com/zjuluolun/BEVPlace2}{https://github.com/zjuluolun/BEVPlace2}.}
\end{abstract}


\begin{figure}[t]
	\begin{center} 
		\includegraphics [width=3.5in]{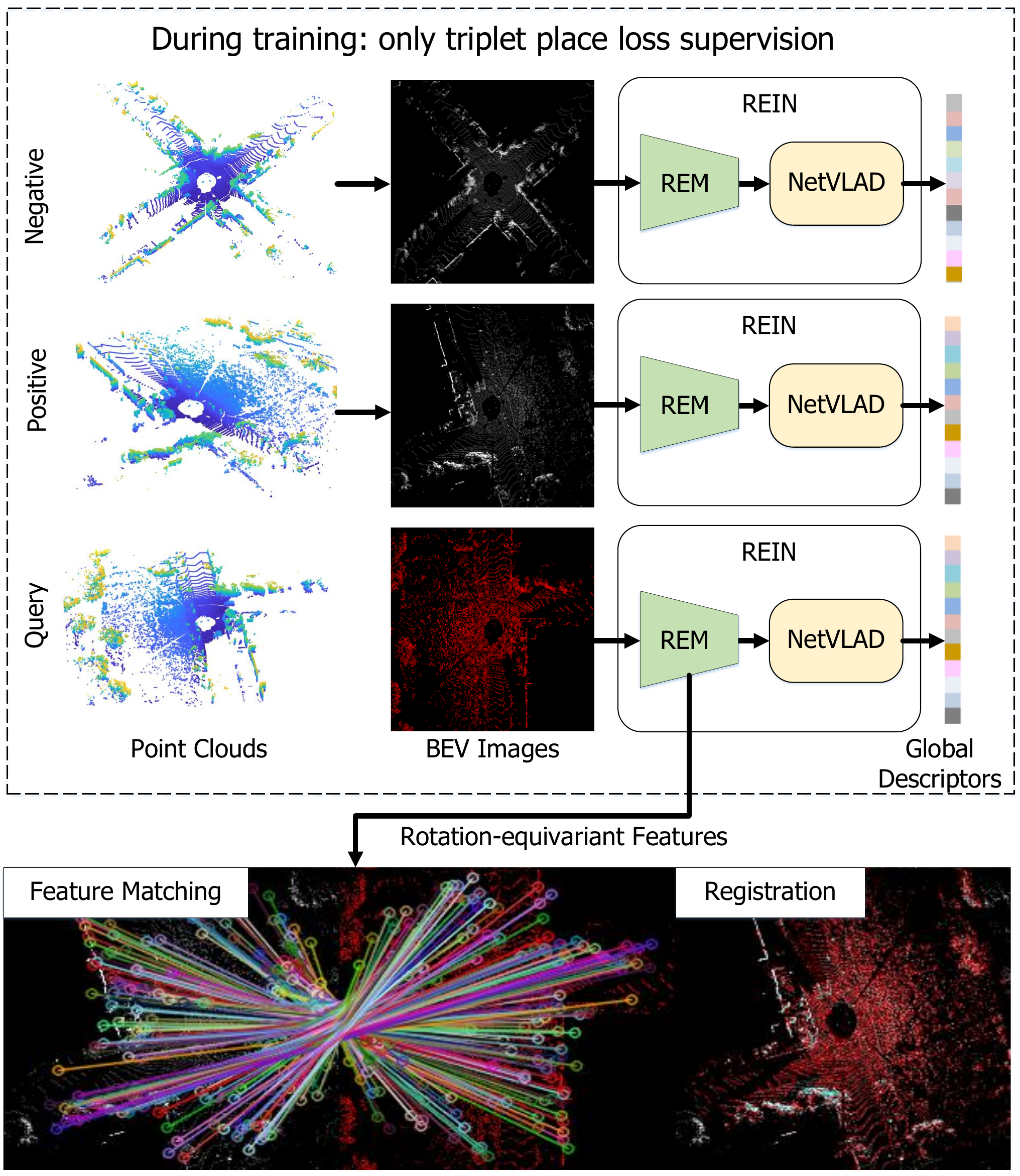}  
            \captionsetup{aboveskip=0pt, belowskip=0pt}
		\caption{The overview of BEVPlace++. It projects point clouds into BEV images, and extracts rotation-equivariant local features and rotation-invariant global descriptors with our devised Rotation Equivariant and Invariant Network (REIN). It performs place recognition with global descriptors and then estimates poses using local features. BEVPlace++ only uses the triplet loss with positive and negative place samples for supervision, while achieving high performance in both place recognition and registration. The query BEV is colored red for better visualization.}
		\label{fig: training}
	\end{center} 
        \vspace{-0.6cm}
\end{figure}

%%%%%%%%%%%%%%%%%%%%%%%%%%%%%%%%%%%%%%%%%%%%%%%%%%%%%%%%%%%%%%%%%%%%%%%%%%%%%%%%
\section{Introduction}
\revise{Global localization estimates the global poses of robots on a map without prior information, playing a pivotal role in enabling full autonomy for unmanned ground vehicles and supporting a range of robotic applications. For example, in Simultaneous Localization and Mapping (SLAM)~\cite{cadena2016past,orbslam2,icp3,suma++}, global localization provides loop closure constraints that reduce accumulative drifts and ensure globally consistent maps. It also offers complementary pose estimations crucial for initializing or recovering pose tracking~\cite{lidarloc1,lidarloc2}.}

\revise{A widely adopted global localization paradigm first builds a global map using structure-from-motion or SLAM. Upon receiving a query image or LiDAR scan, the method searches the map for the best-matched place and computes the global pose through sensor data registration. While image-based global localization methods~\cite{nister2006scalable, lazebnik2006beyond, orbslam2,pronobis2006discriminative,williams2009comparison,2012Bags} have been extensively developed, leveraging robust features such as ORB~\cite{orb} and SIFT~\cite{2004Distinctive}, they are sensitive to illumination and view changes due to perspective imaging. In contrast, LiDAR-based methods, benefiting from active depth sensing, are naturally robust to lighting variations~\cite{pointnetvlad} and provide precise depth information for accurate localization. The evolution of LiDAR technology has made sensors more accurate, compact, and cost-effective, making LiDAR-based global localization an important topic in 3D visual SLAM and driving its widespread adoption in mobile robotics and autonomous vehicles.}

\revise{Despite the great potential of LiDAR global localization, it faces significant challenges in generalization ability. The reasons are twofold. First, since LiDAR scanners detect depth under fixed angular resolutions, point clouds are correspondingly dense for nearby scenes but sparse for distant ones, with varying sparsity across different types of LiDAR scanners. This makes it difficult to extract stable keypoints and distinctive features, which can lead to poor localization accuracy. Second, current learning-based global localization methods ~\cite{du2020dh3d, kamorowski2022egonn, lcdnet, lcrnet} usually need pose supervision. As accurate ground truth poses are expensive to obtain, these methods are difficult to train and may hardly generalize to point clouds collected from different types of sensors.}

\revise{This work introduces BEVPlace++, a fast, robust, and lightweight LiDAR global localization method for 3-DoF global pose estimation without knowing the initial pose. To relieve the influence of the sparsity of point clouds,  BEVPlace++ uses Bird's Eye View~(BEV) images as a simple yet effective and lightweight representation. Unlike the broadly used range images \cite{chen2021rangemcl,overlaptransformer,chen2024joint,ma2023cvtnet}, BEV images provide more stable object scales and relationships, offering better generalization and easier deployment across different LiDAR sensors. We further show that CNN features of BEV images are inherently distinctive and robust to the sparsity of LiDAR point clouds, allowing for accurate pose estimation between BEV image pairs without accurate pose supervision. Additionally, we design a rotation equivariant Module~(REM) to enhance BEVPlace++'s robustness to view changes. Both theoretical and statistical validations confirm the rotation equivariance of our devised REM, enabling BEV images with large view changes to be matched. Building upon REM, we propose a Rotation Equivariant and Invariant Network (REIN) by cascading REM with NetVLAD~\cite{arandjelovic2016netvlad} to generate rotation equivariant local features and rotation invariant global descriptors. Unlike previous works requiring accurate pose labels~\cite{du2020dh3d, logg3dnet, lcdnet, lcrnet}, BEVPlace++ achieves both place recognition and pose estimation using only coarse positive/negative place labels for supervision. An overview of the proposed method is depicted in Fig.~\ref{fig: training}. We use triplet supervision to minimize the feature distance of similar BEV images and maximize the distance of dissimilar ones. Benefiting from our designed translation and rotation equivariant network, such place supervision is sufficient to extract distinctive features on BEV images, allowing direct accurate pose estimation with Random Sample Consensus (RANSAC)~\cite{fischler1981random} matching between BEV images.}

In summary, the contributions of this work are four folds:
\revise{\begin{itemize}
    \item We introduce a weakly supervised BEV image-based LiDAR global localization framework named BEVPlace++. Exhaustive experiments on 7 public datasets demonstrate its state-of-the-art performance in place recognition, loop closure, global localization, and SLAM, with strong generalization to diverse environments and sensor types.
    \item We provide an empirical analysis showing that deep CNN features of BEV images are naturally discriminative for local feature matching under translation motions. Based on this observation, we propose a Rotation Equivariant Module (REM) to preserve the distinctiveness of deep features under rotations by directly leveraging CNN backbones within the module.
    \item We propose a novel Rotation Equivariant and Invariant Network (REIN) combining REM and NetVLAD. REIN makes BEVPlace++ robust to yaw angle view changes by extracting rotation-invariant global descriptors for place recognition and local rotation-equivariant descriptors for pose estimation. Additionally, REIN does not require accurate pose supervision, relying instead on coarse place information for training.
    \item We heve open-sourced BEVPlace++ contributing to the community at \url{https://github.com/zjuluolun/BEVPlace2}.
\end{itemize}
}

 BEVPlace++ is an extension of our previous conference paper BEVPlace~\cite{bevplace}. BEVPlace uses group convolution to extract rotation equivariant local features from BEV images and achieves place recognition by global descriptor matching. 
 Compared to BEVPlace, BEVPlace++ extends in four critical aspects: 1)~a novel REIN network, with a faster and more light-weight rotation equivariant and invariant feature encoder; 2)~a deeper analysis of the rotation and translation equivariance of CNN features; 3)~a pose estimator for achieving 3-DoF poses global localization; 4)~more extensive and comprehensive experimental evaluations on various datasets, demonstrating superior performance in place recognition, loop closure detection, and global localization.

%%%%%%%%%%%%%%%%%%%%%%%%%%%%%%%%%%%%%%%%%%%%%%%%%%%%%%%%%%%%%%%%%%%%%%%%%%%%%%%%
\section{Related Work} \label{sec:related_work}
This section briefly overviews LiDAR-based global localization methods in the literature. 
\revise{Following existing surveys~\cite{yin2023survey,luo20243dlprsurvey}, we categorize related works into four groups based on the integration degree of place recognition and pose estimation: 1) place recognition-only, that retrieves the most similar place using global descriptors; 2) global pose estimation, that directly estimates the global poses without retrieving places; and 3) place recognition followed by local pose estimation, that ﬁrst achieves place recognition and then estimates the robot pose via a pose estimator. 4) As using BEV image representation for LiDAR localization, we additionally review the image processing techniques for BEV images.}

\textbf{1) Place recognition-only.} \revise{These methods generate global descriptors for LiDAR scans and perform place recognition through descriptor retrieval.} The pose of the retrieved place is then considered as the estimated pose. The key to place recognition is creating global descriptors that ensure similar scans are close in descriptor space while dissimilar ones are far apart. Early methods usually exploit the point statistics to represent the point cloud appearance. For example, M2DP~\cite{m2dp} projects a point cloud to multiple 2D planes and generates a density signature for points in each plane. The singular value decomposition (SVD) components of the signature are then used to compute a global descriptor. Scan Context~\cite{kim2018scan,kim2021scan} partitions the ground space into bins according to both azimuthal and radial directions and encodes the 2.5-D information within an image. 

Recently, learning-based place recognition has been a hot topic, and many methods have been proposed. PointNetVLAD~\cite{pointnetvlad} leverages a network to project each point into a higher dimension feature and then uses NetVLAD~\cite{arandjelovic2016netvlad} to generate global features. To take advantage of more contextual information, PCAN~\cite{pcan} introduces the point contextual attention network that learns attention to task-relevant features. Both PointNetVLAD and PCAN cannot capture local geometric structures due to the independent treatment for each point. Thus, the following methods focus on extracting more distinctive local features considering the neighborhood information. LPD-Net~\cite{liu2019lpd} adopts an adaptive local feature module to extract the handcrafted features and uses a graph-based neighborhood aggregation module to discover the spatial distribution of local features. EPC-Net~\cite{epcnet} improves LPD-Net by using a proxy point convolutional neural network. SOE-Net~\cite{soe} introduces a point orientation encoding (PointOE) module. Minkloc3D~\cite{mickloc3d,mickloc3dv2} uses sparse 3D convolutions in local areas and achieves state-of-the-art performance on the benchmark dataset. Recently, some works including SVT-Net~\cite{svtnet}, TransLoc3D~\cite{transloc3d}, NDT-Transformer~\cite{ndtformer}, and PPT-Net~\cite{pptnet} leverage the transformer-based attention mechanism~\cite{attention} to boost place recognition performance. Some methods explore the potential of the range image representation of point clouds. OverlapNet~\cite{overlapnet} uses the overlap of range images to determine whether two point clouds are at the same place and uses a Siamese network to estimate the overlap. OverlapTransformer~\cite{overlaptransformer} further uses a transformer architecture to learn rotation-invariant global features. CVTNet~\cite{ma2023cvtnet} combines range images and BEV images to perform matching. It transforms BEV images into a format similar to range images to achieve rotation-invariance. The range image-based methods usually have better generalization ability but suffer from the scale distortions of translation movements of point clouds. 

\textbf{2) Global pose estimation.} These methods utilize the local characteristics of point clouds through geometric measures to generate local descriptors, which are then directly matched with a global map to determine the global poses. For example, the fast point feature histograms (FPFH)~\cite{rusu2009fast} uses the relationship between the neighbors of interest points and encodes them into a histogram. The Signature of Histograms of OrienTations (SHOT)~\cite{tombari2010unique} builds a local reference frame and leverages the orientation distribution of the normals in local regions. \revise{Although these methods can align local LiDAR scans, they are sensitive to point cloud density and noise.} To leverage the local descriptors more efficiently, the keypoints voting method~\cite{2013Place} performs localization using the 3D Gestalt descriptors. This method relies on robust keypoint extraction, while repeatable 3D keypoint detection is still an open problem in the literature. Some studies~\cite{siftloc,robust2021shan} directly extract conventional image descriptors such as ORB~\cite{orb} and SIFT~\cite{2004Distinctive} from LiDAR images for place recognition. SegMatch~\cite{dube2017segmatch} performs segmentation on LiDAR scans and builds a segment map. During online global localization, it extracts segments from the query LiDAR scan and matches them with the map. The following SegMap~\cite{2018segmap} further assigns the segment descriptors learned by a neural network to improve the matching performance. The main drawback of these methods is their reliance on traveling relatively long distances to extract distinctive segments by gathering multiple scans.

\begin{figure*}[ht]
	\begin{center} 
		\includegraphics [width=7in]{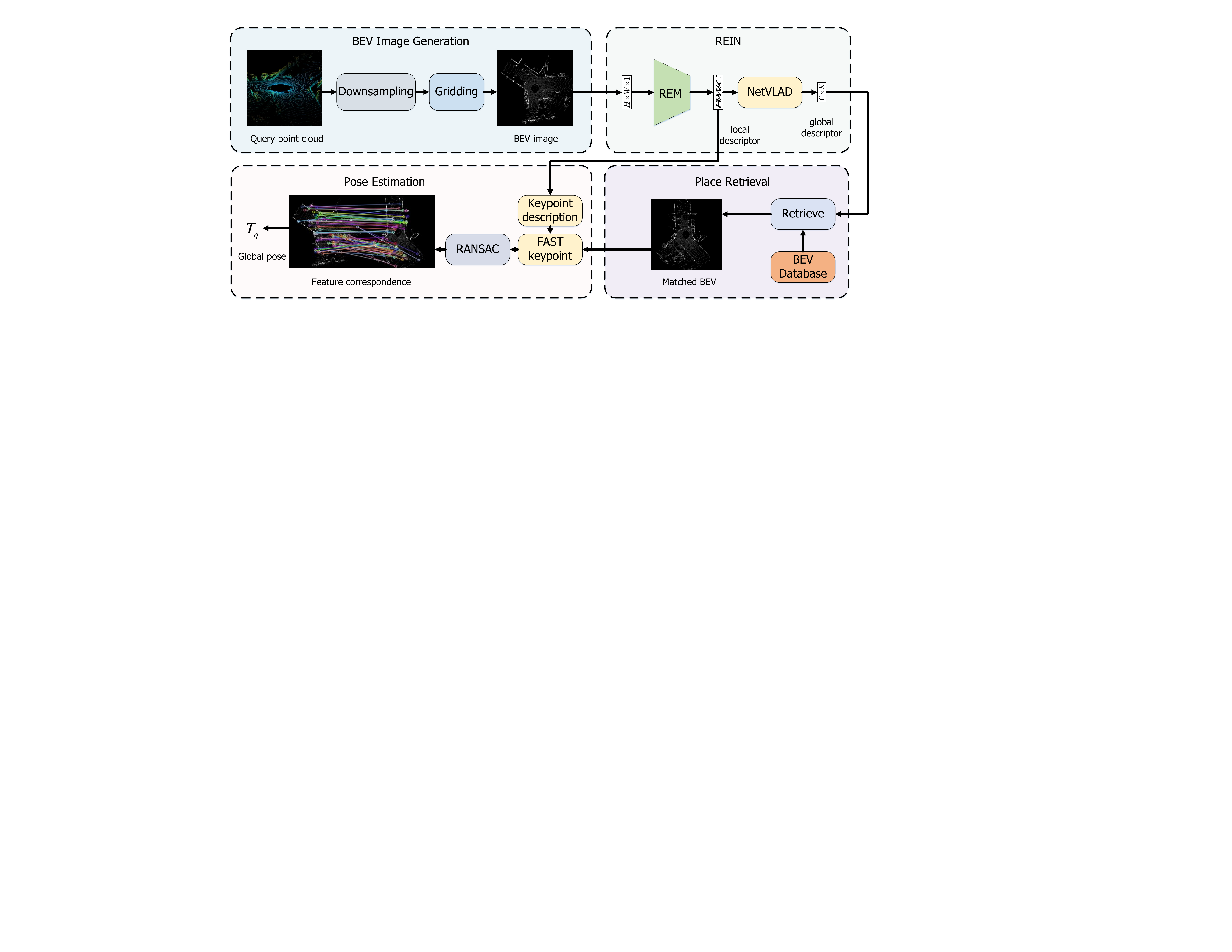}  
            \captionsetup{aboveskip=0pt, belowskip=0pt}
		\caption{The pipeline of BEV image based global localization. Given a query point cloud, we generate its BEV image and sequentially extract local and global descriptors. The global descriptor is used for retrieving the most similar BEV from a pre-built database. The local descriptors are reused for pose estimation with RANSAC. The global pose of the query is computed as the combination of the pose of the match BEV and the relative pose.}
		\label{fig:pipeline}
	\end{center} 
    \vspace{-0.6cm}
\end{figure*}

\textbf{3) Place recognition followed by local pose estimation}. These approaches have been commonly used in visual global localization. However, it is not widely adopted in LiDAR global localization because few LiDAR local features have reached the same level of maturity as visual features like ORB and SIFT. To tackle the problem, BVMatch~\cite{bvmtach} projects point clouds into BEV images and extracts handcrafted BVFT features from the images. It then uses the bag-of-words model~\cite{nister2006scalable, lazebnik2006beyond, 2012Bags} to generate global features. BoW3D~\cite{bow3d} extracts local features directly from sparse point clouds. It also adopts the bag-of-words model to generate global descriptors and is expected to show moderate generalization ability. Another line of work adopts deep learning techniques. For example, DH3D~\cite{du2020dh3d} uses the 3D local feature encoder and detector to extract local descriptors. It embeds the descriptors to a global feature for place recognition and aligns the matched LiDAR pairs using RANSAC~\cite{fischler1981random}. The following methods such as LCDNet~\cite{lcdnet}, LoGG3D-Net~\cite{logg3dnet}, and LCRNet~\cite{lcrnet} use a similar strategy and unify place recognition and pose estimation into one framework. LCDNet extracts distinctive local features with pointvoxel-RCNN \cite{pvrcnn} and performs data association with the Sinkhorn algorithm \cite{ot}. LoGG3D-Net \cite{logg3dnet} introduces a local consistency loss to guide the network toward learning local features consistently across revisits. LCRNet \cite{lcrnet} exploits novel feature extraction and pose-aware attention mechanism to precisely estimate similarities and 6-DoF poses between pairs of LiDAR scans. By jointly learning local and global descriptors, these methods show satisfactory global localization performance. However, they may not adapt well when the point clouds are out of the distribution of the training data. BTC~\cite{btc} extracts the keypoints of a point cloud by projecting the points to planes. It generates triangle descriptors and develops an efficient matching strategy. 

\revise{\textbf{4) BEV-based methods.} The primary challenge in processing BEV images is designing rotation-robust and discriminative features. Although some handcrafted features such as BVFT~\cite{bvmtach}, HOPN~\cite{hopn}, and RING~\cite{xu2023ringplus} have shown promising performance, learning BEV features with neural networks remains rarely studied. Some BEV-based end-to-end networks~\cite{liu2023bevfusion,li2023powerbev} work well for segmentation, but they do not extract features for localization. In this work, we propose the rotation equivariant module (REM) to address this gap. Notably, rotation equivariant networks have been used in other vision tasks such as object detection~\cite{FRED, redet} and image matching~\cite{lref}. Compared to the rotation equivariance design in these methods and previous BEVPlace~\cite{bevplace}, our REM is explicitly designed to preserve the distinctiveness of deep features by directly leveraging CNN backbones within the architecture. This enables REM to capture richer and more robust features in a rotationally invariant manner.}

%%%%%%%%%%%%%%%%%%%%%%%%%%%%%%%%%%%%%%%%%%%%%%%%%%%%%%%%%%%%%%%%%%%%%%%%%%%%%%%%
\section{BEV-based Global Localization Pipeline}
\label{sec:BEV-localization}
Our method uses bird's eye view (BEV) images as an intermediate representation of LiDAR data to perform global localization. As shown in Fig.~\ref{fig:pipeline}, we project point clouds into BEV views and extract distinct local and global descriptors with the rotation equivariant and invariant network (REIN). Then, we employ a two-stage localization paradigm, i.e., place recognition followed by pose estimation. In place recognition, we perform place recognition with the global descriptors and find the top-matched BEV image from a pre-built database. In pose estimation, we exploit the local descriptors to estimate the relative pose between the query and matched BEV image pair using RANSAC. The global pose of the query is finally computed as the combination of the stored pose of the matched frame and the relative pose.
In the following, we first explain the BEV image generation process, and then briefly describe the proposed REIN network. Finally, we elaborate on our proposed two-stage localization inference pipeline.

\subsection{Bird's Eye View Representation}
\revise{Following existing 3-DoF localization works~\cite{kim2021scan,chen2021rangemcl,xu2023ringplus}, we assume that when a ground vehicle moves within a small local area, it travels on a surface that is mostly flat.} Based on this assumption, we generate BEV images through the orthogonal projection and concentrate on estimating the pose in 3-DoF, including (x, y, yaw). The (z, pitch, roll) can also be derived from the stored pose of the matched frame, but they are not the primary focus of this article. The BEV image representation shows benefits in many aspects of our hierarchical localization system. In place recognition, BEV images offer a comprehensive view of the distribution of road elements. Therefore, it is more intuitive and stable to extract global descriptors from such images to depict the structural information of the scene. In the context of pose estimation, the transformations of BEV image pairs are estimated by solving the BEV image matching problem, which is fast because of the lightweight BEV representation. The lightweight BEV representation also benefits the storage resource which is important for real-world applications and multi-robot communication.

We use the normalized point density to construct BEV images \cite{bvmtach}. Let $\mathcal{P}=\{P_i|i=1,...,N_p\}$ represent a point cloud formed by LiDAR points $P_i=(x_i,y_i,z_i)$ with a total of $N_p$ points. We use the right-hand Cartesian coordinate system, where the x-axis points to the right, the y-axis points forward, the z-axis points upward, and the x-y plane is the ground plane. For a point cloud $\mathcal{P}$, we first use a voxel grid filter with the leaf size of $g$ meters to evenly distribute the points. Then we discretize the ground plane into grids with a resolution of $g$ meters. Considering a [$-D $ m $, D $ m] cubic window centered at the coordinate origin, BEV image $\textbf{I}(u,v)$ can be considered as a matrix of size $\lceil{\frac{2D}{g}} \rceil \times \lceil{\frac{2D}{g}}\rceil$. The BEV pixel value $\textbf{I}(u,v)$ is computed as the normalized point density with
\begin{equation}
        \label{eq:BV image}
        \begin{aligned}
        \textbf{I}(u,v)=\frac{\min(N_g,N_m)}{N_m},
        \end{aligned}
\end{equation}
where $N_g$ denotes the number of points in the grid at position $(u,v)$ and $N_m$ the normalization factor. $N_m$ is set as the maximum value of the point cloud density.

Unlike traditional BEV image, also called elevation map~\cite{kim2018scan,kim2021scan}, which stores the maximum height of the points in each bin, we use the point densities. \revise{This is because the elevation map is influenced by the sensor's orientation, as the maximum measured height changes with the distance between the scanner and the object. In contrast, the point density on a surface is less sensitive to viewpoint changes~\cite{bvmtach,mapclose}.}

\subsection{Rotation Equivariant and Invariant Netwrok}
We propose a novel REIN network to extract local features of BEV images through devised rotation equivariant module (REM) and invariant global descriptors using NetVLAD~\cite{arandjelovic2016netvlad}. Given a query BEV image $\textbf{I}_q\in \mathbb{R}^{H\times W}$,  REM produces a feature map $\textbf{F}\in \mathbb{R}^{{H}'\times {W}'\times C}$, where $H', W', C$ is the height, width, and feature channels. Since the output feature map of REM is downsampled compared to the raw image size, we upsample the feature map with bilinear interpolation to assign descriptors for keypoints detected in the BEV image conveniently. These local descriptors are first aggregated by NetVLAD to generate a global descriptor $\textbf{V}\in\mathbb{R}^{K\times C}$ for place recognition where $K$ is the number of clusters in NetVLAD. The local descriptors are also reused in pose estimation. We will introduce the design of REIN in Sec.~\ref{sec: bevplace++}.

\subsection{Place Recognition}
Place recognition assumes that point clouds close in feature space are also close geographically. We can retrieve a most apparently similar frame to a query BEV image $\textbf{I}_q$ from a pre-built BEV database according to the distances between global descriptors. The stored prior pose $\textbf{T}_m$ of the matched frame can be regarded as a coarse estimation of the query pose.

\textbf{1) BEV Database construction.} The BEV database contains necessary map information to achieve global localization, including keyframe BEV images, their associated global poses, and their descriptors. Suppose that the vehicle mounting a LiDAR sensor traverses a specific working area and collects LiDAR scans along the way. Every collected LiDAR scan in this traversal is tagged with a global pose by building a map using SLAM or GPS information. We generate a global descriptor for each collected LiDAR keyframe using our BEVPlace++ model. We denote the database formed by $n$ keyframes as a set
\begin{equation}
        \label{recover}
        \begin{aligned}
                \mathcal{D}=\{(\mathbf{I}_i,\mathbf{T}_i, \mathbf{V}_i)\}_{i=1,2,...,n}
        \end{aligned}
\end{equation}
where $\mathbf{T}_i\in\mathbb{R}^{3\times 3}$ and $\mathbf{V}_i\in\mathbb{R}^{K\times C}$ are the corresponding global pose and global descriptor of the BEV image $\mathbf{I}_i$. The database construction is performed online or offline according to specific tasks. For localization within a global map, the database is typically constructed offline during the map-building process, far from the time of current application. \revise{For loop closure in SLAM, the database is constructed online, as loop closure operates as a SLAM module designed to detect loops and correct accumulated drift in real time.} 

\textbf{2) Place recognition.} For the query BEV image $\textbf{I}_q$, we generate its global descriptor $\textbf{V}_q$ with BEVPlace++ feature extractor. The matched frame is computed through the nearest global descriptor searching:
\begin{equation}
    \label{eq: rotation_equivariance}
    \begin{aligned}
        m = \mathop{\arg}\limits_{i=1,2,...,n}\min ||\mathbf{V}_q-\mathbf{V}_i||_2.
    \end{aligned}
\end{equation}

We regard the associated pose $\mathbf{T}_m$ as a rough estimation of the query pose $\mathbf{T}_q$. In practice, we use PCA to reduce all the global descriptors into 512-dim to speed up the reference.

\subsection{Pose Estimation} 
After finding the matched pair of BEV images, we match two BEV images by local descriptor matching and compute the BEV pose with RANSAC~\cite{fischler1981random}. 

\textbf{1) BEV image matching.} We first extract FAST~\cite{rosten2006machine} keypoints from the BEV images to enable fast and accurate keypoint detection. Furthermore, on BEV images, the detected FAST keypoints usually have good repeatability since they are usually located at objects with vertical structures in the environment (\emph{e.g.}, poles, facades, guideposts). \revise{Moreover, the FAST detector requires no training, making it suitable for the weakly supervised design of our framework.} We then assign each keypoint with a local descriptor interpolated from the REM feature map. We perform local feature matching between the BEV image pair and use RANSAC to estimate the relative transform with the keypoint correspondences. 

\textbf{2) Global pose recovery.} Since BEV images are generated from point clouds with orthogonal projection, the transform between BEV images is rigid. Once we know the transformation of BEV images, we can recover the transformation between the corresponding point clouds through a similar transformation. After obtaining the transform of $(\textbf{I}_q(u,v),\textbf{I}_m(u,v))$, we have:
\begin{equation}
        \begin{aligned}
                &\textbf{I}_q(u,v)=\textbf{I}_m(u',v') \\
                &u'=\cos(\theta)u+\sin(\theta)v+t_u \\
                &v'=-\sin(\theta)u+\cos(\theta)v+t_v,
        \end{aligned}
\end{equation} 
where $(t_u,t_v,\theta)$ are transform parameters. The transform matrix $\textbf{T}_{mq}$ of the corresponding scan+ pair ($\mathcal{P}_m, \mathcal{P}_q)$ is 
\begin{equation}
        \label{recover}
        \begin{aligned}
                \textbf{T}_{mq}=  
                \begin{pmatrix} 
                        \cos(\theta) & \sin(\theta) &  g{t_u} \\
                        -\sin(\theta) & \cos(\theta)  & {g}{t_v} \\
                        0 & 0 & 1  \\
                        % 0 & 0 & 0 & 1
                \end{pmatrix}
        \end{aligned}
\end{equation}
where $g$ is the BEV image resolution. As the global pose $\textbf{T}_m$ of the matched image is stored in the database, we could obtain the global pose of $\mathcal{P}_q$ as
\begin{equation}
        \label{recover}
        \begin{aligned}
                \textbf{T}_q = \textbf{T}_m\textbf{T}_{mq}
        \end{aligned}
\end{equation}

%%%%%%%%%%%%%%%%%%%%%%%%%%%%%%%%%%%%%%%%%%%%%%%%%%%%%%%%%%%%%%%%%%%%%%%%%%%%%%%%
\section{Rotation Equivariant and Invariant Network}
\label{sec: bevplace++}

This section details the design of our rotation equivariant and invariant network (REIN), which includes a feature encoder for generating local features and a pooling layer for aggregating these local features into global descriptors. We begin by highlighting our finding that modern convolutional networks can effectively serve as distinctive feature encoders for BEV images under translation displacements. Then, we introduce a novel Rotation Equivariant Module (REM) to extract local features to achieve robustness to rotational view changes. Building on the rotation-equivariant local features, we show that our global descriptor, pooled by NetVLAD, is rotation-invariant. Finally, we elaborate on the network training strategy. 

\subsection{BEV Features Generation Through CNNs}
Here, we provide a detailed statistical analysis of our finding that modern CNNs can serve as distinctive feature extractors for BEV images with translational movements. For clarity, we denote the process of feature extraction as
\begin{equation}
        \label{equ: feature_extraction}
        \begin{aligned}
                \textbf{F}=\phi(\textbf{I}),
        \end{aligned}
\end{equation}
where features $\textbf{F}\in \mathbb{R}^{H\times W\times C}$ are derived from a BEV image $\textbf{I}\in \mathbb{R}^{H\times W}$ through a feature extractor $\phi$. 
% We mainly focus on the keypoint features for efficiency. 
We denote the feature of a keypoint located at the coordinate $(u,v)$ as $\textbf{f}_{u,v}\in \mathbb{R}^{C}$, which also represents the element of $\textbf{F}$ located at u-th row and v-th column with $C$ channels.

\begin{figure}[t]
	\begin{center} 
		\includegraphics [width=3.4in]{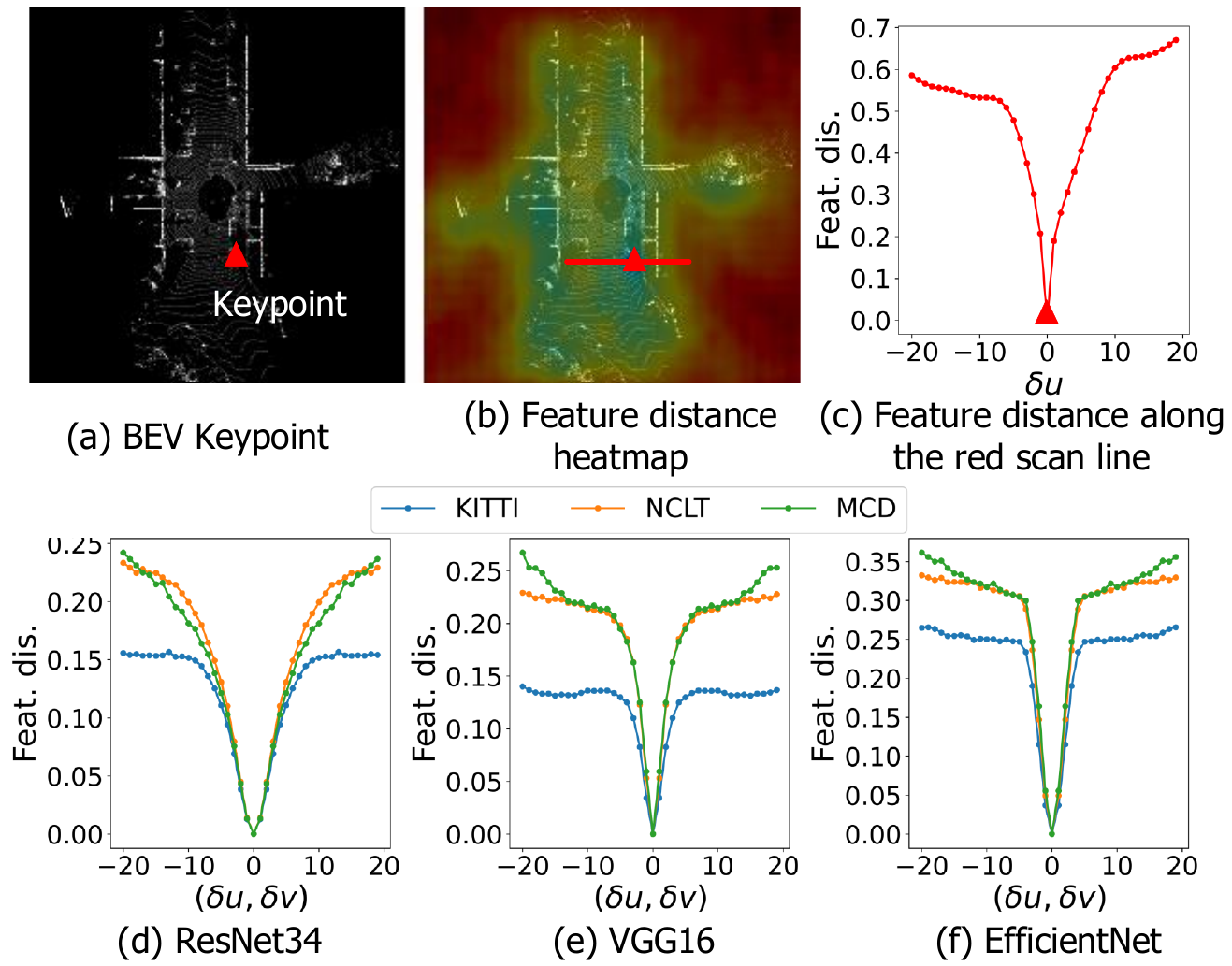}  %[width=0.25\textwidth]
            \captionsetup{aboveskip=0pt, belowskip=0pt}
		\caption{ Statistical analysis: BEV feature distance distribution with respect to translation displacements with different CNN architectures and datasets. (a) A keypoint on a BEV image highlighted by a red triangle. (b) The feature distance heat map relative to all other pixels. (c) The feature distance variation along the red scan line in (b). (d) The average feature distance distribution concerning translation displacements (with $\delta u = \delta v$) using ResNet34 as the feature extractor. (e) The average feature distance distribution using VGG16. (f) The average feature distance distribution using EfficientNet.}
		\label{fig: feature_distribtuion}
	\end{center}  
    \vspace{-0.6cm}
\end{figure}

\textbf{1) Distinctive BEV features from CNNs.} We investigate the feature variation to validate the distinctiveness of BEV features from CNN extractors. For each keypoint feature $\textbf{f}_{u,v}$ extracted on our devised LiDAR BEV image, we compute its Eucleadian distance to its neighbor features by
\begin{equation}
        \label{equ: translation_equ_feature}
        \begin{aligned}
                d(\textbf{f}_{u,v}, \textbf{f}_{u+\delta u,v+\delta v}) = ||\textbf{f}_{u,v} - \textbf{f}_{u+\delta u,v+\delta v}||_2,
        \end{aligned}
\end{equation}
where $\delta u,\delta v$ are the translation displacements. Fig.~\ref{fig: feature_distribtuion} (a), (b), and (c) demonstrate an example of feature distance distribution for a specific feature extracted from a popular CNN, ResNet~\cite{resnet} with its pretrained model\footnote{https://pytorch.org/vision/stable/models.html} provided by PyTorch. We compute the feature distance of the keypoint feature shown in (a) relative to all other pixels and present the feature distance heatmap in (b). Additionally, (c) displays the numerical feature distance variation along the red scan line in (b). As can be seen, even without fine-tuning, the BEV feature distance increases as the translation displacement grows, indicating its inherent distinctiveness.

We further plot the average feature distance with respect to translation displacements using different CNN backbones, including ResNet34~\cite{resnet}, VGG16~\cite{vgg} and EfficientNet~\cite{tan2019efficientnet}, on different datasets, such as KITTI~\cite{2012KITTI}, NCLT~\cite{2016Universitynclt}, and MCD~\cite{MCD}, shown in Fig.~\ref{fig: feature_distribtuion} (e), (f), and (g). These datasets include point clouds collected by various types of LiDAR scanners with different fields of view and sparsity levels across diverse environments. As can be seen, spatially close features exhibit small feature distances while distant ones show larger feature distances, regardless of the dataset setup or used CNN backbones, which could be formulated as 
\begin{equation}
        \label{equ: translation_distinctive}
        \begin{aligned}
                d(\textbf{f}_{u,v}, \textbf{f}_{u+\delta u_1,v+\delta v_1}) < d(\textbf{f}_{u,v}, \textbf{f}_{u+\delta u_2,v+\delta v_2}), \\
                for \quad ||(\delta u_1,\delta v_1)||<||(\delta u_2,\delta v_2)||.
        \end{aligned}
\end{equation}
\revise{This inherent distinctiveness reveals that geometrically adjacent BEV points maintain adjacency in the feature space, highlighting the discriminative capability of the feature encoder and its ability to preserve spatial relationships.} Such properties are crucial for accurately estimating pose, as they help in distinguishing between different keypoints and ensuring that their spatial relationships are preserved.

\textbf{2) Matching with distinctive features.} CNNs demonstrate translation equivariance \cite{groupconv} (up to edge-effects) due to local connectivity and weight sharing inherent in convolution operations. Given a BEV image $\textbf{I}'$ obtained from $\textbf{I}$ with translational motion, the keypoint feature $\textbf{f}'_{u',v'}$ in $\textbf{I}'$ should ideally be identical to its positional corresponding feature $\textbf{f}_{u,v}$ in $\textbf{I}$. Such correspondence should be unique according to Eq.~\ref{equ: translation_distinctive}, that is  
\begin{equation}
        \label{equ: translation_distinctive_quivatiant}
        \begin{aligned}
                d(\textbf{f}'_{u',v'}, \textbf{f}_{u,v}) < d(\textbf{f}'_{u',v'}, \textbf{f}_{u+\delta u,v+\delta v}), \\
                for \quad ||(\delta u,\delta v)||>0.
        \end{aligned}
\end{equation}
Leveraging this characteristic, we establish BEV feature correspondences through nearest neighbor search and utilize RANSAC to solve for pose estimation.

\begin{figure}[t]
	\begin{center} 
		\includegraphics [width=3.4in]{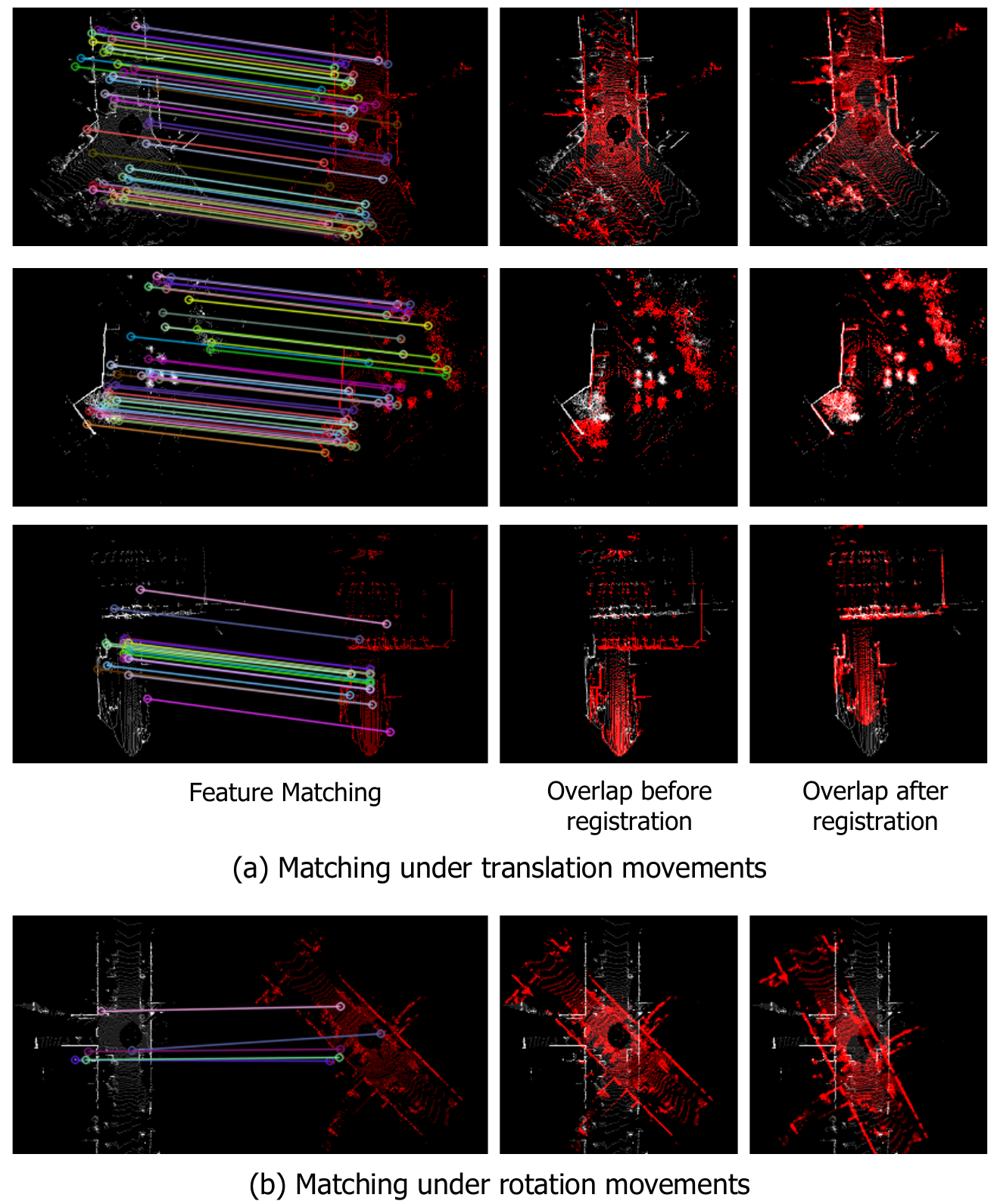} 
            \captionsetup{aboveskip=0pt, belowskip=0pt}
		\caption{RANSAC matching utilizes BEV features extracted from a randomly initialized ResNet34. (a) Matching under translations. The three rows correspond to BEV images from the KITTI, NCLT, and MCD datasets, respectively. We illustrate the feature correspondences after RANSAC, BEV overlap before registration, and BEV overlap after registration from left to right. (b) Matching fails under rotations.}
		\label{fig: feature_distribtuion_matching}
	\end{center} 
    \vspace{-0.7cm}
\end{figure}

Fig.~\ref{fig: feature_distribtuion_matching} (a) illustrates feature matching results using ResNet34 under large translations. Each row corresponds to BEV images from the KITTI, NCLT, and MCD datasets, respectively. Following pose estimation, the overlaid BEV image pairs indicate our method's capability to accurately estimate significant translation movements. \revise{It is noted that feature matching could fail under rotations, as shown in Fig.~\ref{fig: feature_distribtuion_matching} (b), since traditional CNNs are not rotation-equivariant and the feature distinctiveness is only preserved under translation transformations.}

\subsection{Rotation Equivariant Module}
\label{sec:ren}
As discussed above, traditional CNNs are translation equivariant but cannot handle rotations well. To achieve more robust pose estimations, we further design a network to extract rotation-equivariant feature maps from BEV images. We introduce a simple and effective rotation equivariant module (REM) to extract rotation equivariant local features. REM uses modern CNNs as basic modules to keep their representation ability to BEV images and achieves rotation equivariance by using the maximum CNN feature response of BEV images under different rotation transforms. It is noted that there are some rotation equivariant networks that have been used in vision tasks such as object detection~\cite{redet,FRED}. Our REM differs from them since REM uses CNN backbone as basic modules and thus is explicitly designed to retain the descriptive capability of CNN backbones to BEV images.

\begin{figure}[t]
	\begin{center} 
		\includegraphics [width=3.4in]{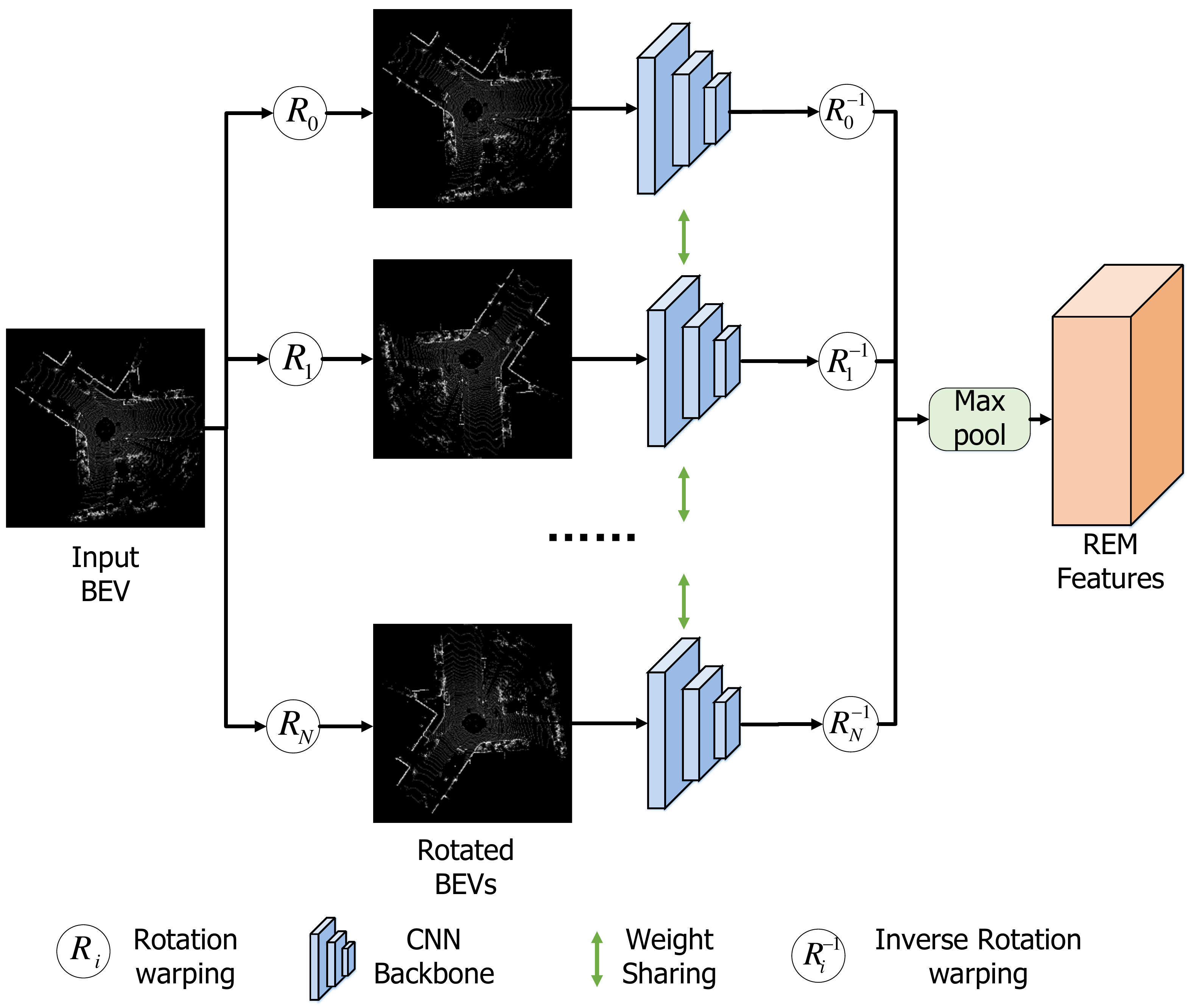}  %[width=0.25\textwidth]
            \captionsetup{aboveskip=0pt, belowskip=0pt}
		\caption{The architecture of rotation equivariant module (REM). Given an input BEV image, REM first warps the image with several rotation transforms. It then generates features for each warped image with weight-shared residual convolutional blocks. The output features are warped back with inverse rotations and are max pooled to obtain a rotation equivariant feature map.}
		\label{fig: ren}
	\end{center} 
    \vspace{-0.6cm}
\end{figure}

\textbf{1) Rotation Equivariance Module Design.} The architecture of REM is as illustrated in Fig.~\ref{fig: ren}. Given an input BEV $\textbf{I}$, we warp it with the $N$ rotation angles from the angle set $\mathcal{R}=\{0,\frac{2\pi}{N_R}, ...,(N-1)\frac{2\pi}{N_R} \}$. For each rotated image, we extract local features with shared residual convolutional modules and rotate the feature map back. The equivariant local features are obtained by performing max pooling between $N$ rotated feature maps. The rotation equivariant features $F$ are obtained by
\begin{equation}
	\label{eq: rotation_invriance}
	\begin{aligned}
		\textbf{F} = \max_{r\in \mathcal{R}} \textbf{R}^{-1}_r\circ{\phi(\textbf{R}_r\circ \textbf{I})}.
	\end{aligned}
\end{equation}
where $\phi$ is the residual convolutional modules.

\begin{figure*}[htbp]
	\begin{center} 
		\includegraphics [width=6in]{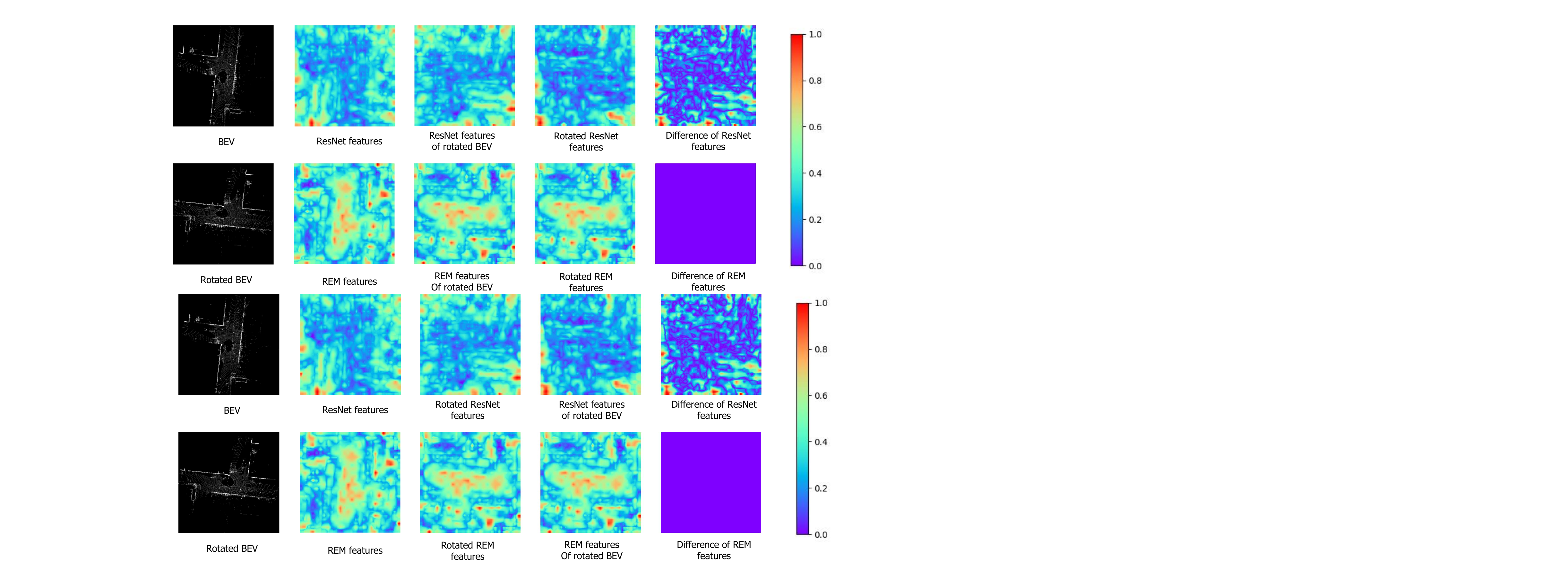}  %[width=0.25\textwidth]
            \captionsetup{aboveskip=0pt, belowskip=0pt}
		\caption{\revisesec{Comparison between ResNet and REM features with rotations. Features are extracted from a pair of rotated BEV images with ResNet and REM. The difference between the ResNet features of the rotated BEV and the rotated ResNet features of the original BEV is large. On the other hand, REM shows rotation equivariance, and the difference between the REM features of the rotated BEV and the rotated REM features of the original BEV tends to be zero.}}
		\label{fig: random_match}
	\end{center} 
    \vspace{-0.6cm}
\end{figure*}

\revisesec{\textbf{2) Rotation Equivariance Analysis.} Applying a rotation $\textbf{R}_\alpha\in\mathcal{R}$ to the BEV image $\textbf{I}$, the output} \revisesec{feature map $\textbf{F}'$ is \begin{equation}
	\label{eq: rotation_invriance_step1}
	\begin{aligned}
		\textbf{F}' & = \max_{r\in \mathcal{R}} \textbf{R}^{-1}_r\circ{\phi((\textbf{R}_r\textbf{R}_\alpha)\circ \textbf{I})}. \\
        & = \max_{r\in \mathcal{R}} \textbf{R}_\alpha \textbf{R}_\alpha^{-1}\textbf{R}^{-1}_r\circ{\phi((\textbf{R}_r\textbf{R}_\alpha)\circ \textbf{I})}\\
        & = \max_{r\in \mathcal{R}} \textbf{R}_\alpha \textbf{R}_{r+\alpha}^{-1}\circ{\phi((\textbf{R}_{r+\alpha})\circ \textbf{I})} 
	\end{aligned}
\end{equation}
Since the rotation group $\mathcal{R}$ is closed under addition, the mapping $r \mapsto r+\alpha$ is a bijection on $\mathcal{R}$. Thus, the set $\{\textbf{R}_{r+\alpha}^{-1}\circ \phi(\textbf{R}_{r+\alpha}\circ \textbf{I})|r\in \mathcal{R}\}$ is identical to the set $\{\textbf{R}_r^{-1}\circ \phi(\textbf{R}_r\circ \textbf{I})|r\in \mathcal{R}\}$. Applying a left multiplication by an additional rotation $R_\alpha\in\mathcal{R}$ to each element of the set does not alter the identity. Thus, we can rewrite $\textbf{F}'$ as: 
\begin{equation}
	\label{eq: rotation_invriance}
	\begin{aligned}
		\textbf{F}' & = \max_{r\in \mathcal{R}} \textbf{R}_\alpha \textbf{R}_{r+\alpha}^{-1}\circ{\phi((\textbf{R}_{r+\alpha})\circ \textbf{I})} \\
        & = \max_{r\in \mathcal{R}} \textbf{R}_\alpha \textbf{R}_{r}^{-1}\circ{\phi(\textbf{R}_{r}\circ \textbf{I})} .
  % & = \textbf{R}_\alpha\circ\textbf{F} .
	\end{aligned}
\end{equation}
The maximum of a set is determined solely by the largest element within it. Therefore, the result remains the same regardless of the order of the elements in the set.  Additionally, rotation operations only induce spatial permutations without altering the values (ignoring edge effects), we have:
\begin{equation}
	\label{eq: rotation_invriance}
	\begin{aligned}
		\textbf{F}' & = \max_{r\in \mathcal{R}} \textbf{R}_\alpha \textbf{R}_{r}^{-1}\circ{\phi(\textbf{R}_{r}\circ \textbf{I})}  \\
         & = \textbf{R}_\alpha\circ \max_{r\in \mathcal{R}}  \textbf{R}_{r}^{-1}\circ{\phi(\textbf{R}_{r}\circ \textbf{I})} \\
         & = \textbf{R}_\alpha\circ \textbf{F}.
  % & = \textbf{R}_\alpha\circ\textbf{F} .
	\end{aligned}
\end{equation}
}This derivation demonstrates that the output features $\textbf{F}$ are rotation-equivariant, as the effect of a rotation $\textbf{R}_\alpha$ on the input image $\textbf{I}$ results in the same rotation applied to the resulting feature map $\textbf{F}$. Theoretically, we need to sample infinite angles to achieve continuous rotation equivariance. In our experiments, we find that a small $N$ can achieve satisfactory performance. This is possible because the pooling and downsample operations in CNN could provide some robustness to small movements. 

Fig. \ref{fig: random_match} illustrates the comparison between ResNet and REM features under rotations. We extract features from a pair of rotated BEV images with ResNet and REM, respectively. As can be seen, the difference between the REM features of the rotated BEV and the rotated REM features of the original BEV tends to be zero, showing the rotation equivariance of REM. On the other hand, ResNet does not own such equivariance and its feature map difference with rotational changes is large.

\textbf{3) Discussion.} We have demonstrated that modern CNN networks are highly effective for extracting distinctive features from our devised BEV images. Leveraging the strengths of CNN modules, our REM network preserves this distinctiveness while further addressing BEV image matching under significant translational and rotational changes. 
Consequently, BEVPlace++ does not require accurate pose supervision, making it convenient for real-world deployment, where obtaining accurate pose information is often challenging.

% \vspace{-2mm}
\subsection{Rotation Invariant Global Descriptor.}
% \textbf{} 
For robust place recognition, we design a network $\beta$ to extract rotation-equivariant features from BEV images such that any rotation transformation on the input image will result in the same global descriptors, which can be formulated as
\begin{equation}
	\label{eq: rotation_invriance}
	\begin{aligned}
		\beta(\textbf{R}\circ \textbf{I}) = \beta(\textbf{I}).
	\end{aligned}
\end{equation}
We use the cascading of REM and NetVLAD to generate such rotation-invariant global descriptors.

\textbf{1) NetVLAD.} NetVLAD \cite{arandjelovic2016netvlad} is a widely used method for pooling descriptors in image retrieval. It assumes similar structures in the environment produce similar distributions of features and summarizes information about the distributions across an image into a global descriptor. We first constructs K cluster centers denoted as $\{\mathbf{c}_i|\mathbf{c}_i\in \mathbb{R}^C\}_{i=1,...,K}$. Denoting $\{\textbf{f}_i |\textbf{f}_i\in \mathbb{R}^C \}_{i=1,...,{H'W'}}$ be the set of local features flattened from the REM feature map $\textbf{F}$, we generate a global feature $\mathbf{V}=[\mathbf{V}_1, \mathbf{V}_2, ..., \mathbf{V}_K]$ of dimension $C × K$. $\textbf{V}_k$ is the weighted sum of residuals of all local features with respect to the k-th cluster center, namely
\begin{equation}
    \label{eq: rotation_equivariance}
    \begin{aligned}
        \mathbf{V}_k = \sum_i a_k(\mathbf{f}_i)(\mathbf{f}_i-\mathbf{c}_k),
    \end{aligned}
\end{equation}
where $a_k(\mathbf{f}_i)$ the soft-assignment of the feature $\mathbf{f}_i$ for the $k$-th cluster center, computed as 
\begin{equation}
    \label{eq: rotation_equivariance}
    \begin{aligned}
 a_k(\mathbf{f}_i) = \frac{e^{\mathbf{w}_k^{T}\mathbf{f}_i+b_k}}{\sum_{k'}e^{{\mathbf{w}_{k'}^{T}\mathbf{f}_i+b_{k'}}}}
    \end{aligned}
\end{equation}

\textbf{2) Rotation Invariance Analysis.} \revise{Let $\{\mathbf{f}_{i}\}_{i=1,...,{H'W'}}$ and $\{\mathbf{f'}_i\}_{i=1,...,{H'W'}}$ denote the local descriptor sets of the raw and the rotated BEV image. As the REM feature map is rotation equivariant, set $\{\mathbf{f}_i\}_{i=1,...,H'W'}$ and $\{\mathbf{f'}_i\}_{i=1,...,H'W'}$ contain the same elements but differ in their permutation due to the rotation. This indicates a one-to-one correspondence between the two sets, ensuring consistency in their cluster assignments. Since summation $\sum$ is a symmetric operation with respect to permutations, we have
\begin{equation}
    \label{eq: rotation_equivariance}
    \small
    \begin{aligned}
        \mathbf{V}_k' & = a_k(\mathbf{f}_1')(\mathbf{f}_1'-\mathbf{c}_k)+a_k(\mathbf{f}_2')(\mathbf{f}_2'-\mathbf{c}_k)+...+a_k(\mathbf{f}_n')(\mathbf{f}_n'-\mathbf{c}_k)\\
        & = \sum_{i}a_k(\mathbf{f}_i')(\mathbf{f}_i'-\mathbf{c}_k) \\
        & = \sum_{i}a_k(\mathbf{f}_i)(\mathbf{f}_i-\mathbf{c}_k)= \mathbf{V}_k.
    \end{aligned}
\end{equation}
This equality demonstrates that the global descriptors $\mathbf{V}_k$ are invariant to rotation, as they include symmetric operations that are unaffected by the permutation of local descriptors.
}

\begin{table*}[ht]\small
\begin{threeparttable} 
\footnotesize
	\centering
	%	\vspace{-3mm}
        \renewcommand\arraystretch{1.2}
	\renewcommand\tabcolsep{4.5pt}
        \captionsetup{aboveskip=0pt, belowskip=0pt}
	\caption{Evaluation Datasets.} 
	\begin{tabular}{lcccccccccc}
		\hline
		\textbf{Dataset}                   & KITTI~\cite{2012KITTI}     & NCLT ~\cite{2016Universitynclt} &MCD~\cite{MCD} & RobotCar~\cite{robotcar} & Inhouse~\cite{pointnetvlad}  &  Apollo~\cite{l3net} &  Mulran~\cite{mulran} &   Our UGV            \\ \hline
        LiDAR &  \makecell{Velodyne\\ HDL-64E} & \makecell{Velodyne \\ HDL-32E}  &  \makecell{Livox\\ Mid-70} & \makecell{SICK \\LD-MRS 3D} &    \makecell{Velodyne\\ HDL-64} &  \makecell{Velodyne \\HDL-64} &  \makecell{Ouster \\ OS1-64 } &  \makecell{RoboSense \\ Helios-32}\\
		\hline
		FoV (HxV)  & 360.0$^\circ\times$ 26.8$^\circ$       & 360.0$^\circ\times$ 41.3$^\circ$  & 70$^\circ\times$ 70 $^\circ$  & 85.0$^\circ\times$ 3.2$^\circ$ & 360.0$^\circ\times$ 26.8$^\circ$ &  360.0$^\circ\times$ 26.8$^\circ$ &  290.0$^\circ\times$ 45.0$^\circ$ &  360.0$^\circ\times$ 70 $^\circ$   \\\hline
        \# of Seq. & 5 & 7  & 6 & 44 & 15 &  6 &  2 &  1 \\\hline
        Scenes  & Country, Highway & Campus &City & City & Campus &   City, Highway &  City &   Campus  \\\hline
        Cities  & Karlsruhe & Michigan & Singapore & Oxford & Singapore &  California &  Sejong &  Changsha
        \\\hline
		Traj. length   & 5 km  & 10 km& 2km &  10 km & 10 km  &  73 km &  23 km  &   1.5 km        \\\hline
        Time span & Single days & Across 1 year & Several days & Across 1 year & Several days &  Several days &  Several days &  Several days \\\hline
        Tasks & PR \& LC \& GL &  PR  \& LC & PR \& GL  & PR & PR &  GL &  GL &  LC\\
		\hline
	\end{tabular}
    \begin{tablenotes} 
		\item PR, LC, and GL correspond to place recognition, loop closure, and global localization. 
     \end{tablenotes} 
	%	\vspace{-6mm}
 
	\label{tab: evaluation_dataset}
\end{threeparttable} 
    \vspace{-0.6cm}
\end{table*}

\subsection{Network Training}
\textbf{1) Loss function.} We aim to train the BEVPlace++ network such that the geographically close BEV images are close in the feature space, and geographically distant BEV images are far apart. We use the lazy triplet loss~\cite{pointnetvlad} to supervise the global descriptor generation. The lazy triplet loss focuses on maximizing the feature distance between a query and its closest/hardest negative sample in the training set, that is 
\begin{equation}
    \label{eq: loss}
    \begin{aligned}
        \mathcal{L} = \max_j\bigl(\max(m+||\textbf{V}_a-\textbf{V}_p||_2 - ||\textbf{V}_a-\textbf{V}_{nj}||_2,0)\bigr),
    \end{aligned}
\end{equation}
where $\textbf{V}_p, \textbf{V}_{nj}$ are the global descriptors of the positive and j-th negative sample of the query, $\textbf{V}_p$ is the global descriptor of the query BEV image, and $m$ is the constant margin. In our design, we do not supervise the REM network since the deep BEV features from REM are inherently distinctive as discussed before. 

\textbf{2) Training setups.} We assume that the ground vehicle mounting a LiDAR sensor traverses a specific working area and collects LiDAR scans along the way. Every LiDAR scan collected in this traversal is tagged with a global pose from a SLAM method or GPS information. Note that, these global poses are not necessarily to be very accurate, as we only need them to determine the positive and negative samples with a rough distance threshold. We use every collected scan tagged with a global pose as a query frame. For each query frame, its positive samples are the ones within $5$ meters away from itself and its negative samples are the other frames. Then, the training process is to traverse all the queries and perform gradient descent under the supervision of Eq. \ref{eq: loss}. We also adopt the hard mining strategy \cite{shrivastava2016training} following NetVLAD after the first 10 training epochs.

\section{Evaluation Setup} \label{sec:experiment}

\subsection{Datasets}
We use 7 public datasets to thoroughly evaluate the performance of our method across various LiDAR sensor setups and data collected from different cities, including KITTI \cite{2012KITTI}, NCLT \cite{2016Universitynclt}, MCD \cite{MCD}, RobotCar~\cite{robotcar}, In-house~\cite{pointnetvlad}, Apollo~\cite{l3net}, and Mulran~\cite{mulran} datasets.
We evaluate loop closing performance on KITTI and NCLT datasets since their data sequences have large loops. We test global localization on KITTI, NCLT, MCD, Apollo, and Mulran datasets as they have accurate ground truth poses. We evaluate place recognition on all the datasets. Our evaluation datasets cover diverse scenes, including city, countryside, and campus, and are collected in large-scale places under large time spans. The point cloud data are of different sparsity and different fields of view due to the usage of the various types of LiDAR sensors. Tab.~\ref{tab: evaluation_dataset} summarizes the meta information of the datasets. 
The point clouds in these datasets differ greatly, presenting sufficient challenges for evaluating reliable single-shot global localization. \revise{We further deploy our method on a real UGV platform equipped with a RoboSense-32 LiDAR (See Sec.~\ref{application}).} 
We detailed setups of each dataset as follows:

\textbf{1) KITTI~\cite{2012KITTI}.} This dataset contains a large number of point cloud data collected by a Velodyne 64-beam LiDAR. We select the sequences ``\textit{00}'', ``\textit{02}'', ``\textit{05}'', ``\textit{06}'', ``\textit{07}'', and ``\textit{08}'' of the Odometry subset for evaluation since these sequences contain large revisited areas. We split the point clouds of each sequence into database frames and query frames for place recognition, following the widely adopted partition~\cite{bow3d,lcdnet,lcrnet}.
We use the refined poses from semantic KITTI \cite{semanticktii} with a distance threshold to determine if a loop closure exists.

\textbf{2) NCLT~\cite{2016Universitynclt}.} This dataset was created at the University of Michigan North Campus using a Velodyne32-HDL LiDAR sensor. {The dataset provides ground-truth poses based on a large SLAM solution using LiDAR scan matching and high-accuracy RTK-GPS. The sequences of the dataset are collected on varying routes and cover different parts of the campus across a year. We select sequences collected in different seasons for evaluation, including ``\textit{2012-01-15}'', ``\textit{2012-02-04}''   ``\textit{2012-03-17}'', ``\textit{2012-06-15}'', ``\textit{2012-09-28}'', ``\textit{2012-11-16}'', and ``\textit{2013-02-23}'' .} 

\textbf{3) MCD~\cite{MCD}.} This dataset is collected over large-scale campus areas at different seasons. We use its point cloud data from a non-repetitive lidar, Livox MID-30. The clouds have a circular field of view and are quite different from the clouds of rotating-beam LiDAR. For evaluation, we use the sequences collected at NTU including ``\textit{ntu\_day\_01}'', ``\textit{ntu\_day\_02}'', ``\textit{ntu\_day\_10}'', ``\textit{ntu\_night\_04}'', ``\textit{ntu\_night\_08}'', ``\textit{ntu\_night\_13}''.

\textbf{4) RobotCar~\cite{robotcar} and In-house~\cite{pointnetvlad}}. Both datasets are broadly used by the recent place recognition method based on unordered points. The RobotCar dataset was created with a SICK LD-MRS LiDAR by repeatedly visiting a route of 10 km in Oxford. It contains 44 sequences collected on different days across a year. The In-house dataset consists of three scenarios, including a university sector (U.S.), a residential area (R.A.), and a business district (B.D.).  It is constructed from Velodyne-64 LiDAR scans, and each contains 5 sequences. Different from the aforementioned datasets that provide single LiDAR scans, these two datasets provide submaps built from consecutive scans.

\revise{\textbf{5) Apollo~\cite{l3net}}. This dataset is collected through multiple traversals along six different routes in the southern San Francisco Bay Area. It encompasses a diverse range of environments, including residential neighborhoods, urban downtown areas, and highways. The data is captured using a Velodyne HDL 64E rotating LiDAR, which provides a range and number of scanned 3D points comparable to the MulRan dataset. All six sequences are used for evaluation.}
 
\revise{\textbf{6) Mulran~\cite{mulran}}. This dataset contains point clouds collected in South Korea using Ouster OS1-64 rotating LiDAR. It provides accurate ground truth generated by VRS-GPS/INS integrated navigation system and a graph SLAM. Following EgoNN~\cite{kamorowski2022egonn}, we use the longest and most diverse sequences collected in Sejong as the evaluation sequences.}

\subsection{Evaluation Metrics}
We use different metrics to evaluate different tasks. For place recognition, we use the recall at top-1 following~\cite{bevplace,alita,pointnetvlad}. For each query, we find its Top-1 match from the database. According to a distance threshold of 5 meters~\cite{bevplace}, we determine whether the prediction is a true positive (TP), a false negative (FN), or a false positive (FP). The recall rate is defined as the ratio of TP over the actual positives, i.e.
\begin{equation}
	\label{eq: error_r}
	\begin{aligned}
		\text{Recall} = \frac{\text{TP}}{\text{TP}+\text{FN}}. 
	\end{aligned}
\end{equation}

For loop closing with loop closure detection and pose estimation, we use the Precision-Recall (PR) curve, average precision, F1 max score, max recall at 100\% precision, mean translation errors, and mean rotation errors. The PR curve shows both recall rates and precision, where precision is the ratio of true positives (TP) over all predicted positives, i.e.
\begin{equation}
	\label{eq: error_r}
	\begin{aligned}
		\text{Precision} = \frac{\text{TP}}{\text{TP}+\text{FP}}. 
	\end{aligned}
\end{equation}
Similar to the evaluation of place recognition, we compute the nearest descriptor distance for a query and retrieve the top-1 match from the database. By setting different descriptor distance thresholds, we calculate the corresponding precision and recall pair and plot a PR curve. The average precision is the area under the PR curve. The F1 score is 
\begin{equation}
	\label{eq: error_r}
	\begin{aligned}
		\text{F1 score} = 2 \times \frac{\text{Precision}\times\text{Recall}}{\text{Precision}+\text{Recall}}. 
	\end{aligned}
\end{equation}

We obtain the max recall at 100\% precision by traversing all the precision and recall pairs. The mean rotation and translation errors are computed with the pose errors of all the true positive queries.

\revise{For global localization, we compute the localization success rate (SR) under a threshold of $(2 m, 5^\circ)$ following \cite{lcrnet}. All the successful localizations are used to compute the mean translation errors and mean rotation errors.}

\section{Experiments}

\begin{table}[t]\footnotesize
	\centering
	%		\vspace{-3mm}
	%	
%	\vspace{-4mm}
    \renewcommand\arraystretch{1.1}
	\renewcommand\tabcolsep{5pt}
        \captionsetup{aboveskip=0pt, belowskip=0pt}
	\caption{Recall at Top-1 on the KITTI dataset.}% * denotes that pose estimation is integrated into the method.} 
	\begin{tabular}{l|cccccc}
		\hline
		Sequence & 00 & 02 & 05 & 06 & 08 & Mean \\ \hline
        M2DP \cite{kim2018scan} & 92.9 & 69.3 & 80.7 & 94.8 & 34.4 & 74.4 \\
		BoW3D     \cite{liu2019lpd}         &71.4 & 15.5&58.7&91.8 & 57.0 & 58.9 \\
        Logg3D\cite{logg3dnet}  &99.6 & 96.1 & 97.5 & \textbf{100.0} & 93.5 &  97.3  \\
		CVTNet     \cite{ma2023cvtnet}         &98.7 & 87.1 & 93.5 & 97.8 & 83.7 &  92.1\\
     RING++~\cite{xu2023ringplus}   &  94.1 &  67.9 &  88.4 &  95.2 &  59.1 &  80.9 \\
		LCDNet  \cite{mickloc3d}       &99.9 & 97.7 & 95.3 & \textbf{100.0} & 94.4 &  97.4 \\
                        
         LCRNet\cite{lcrnet}  &  98.8 & 91.2 & 99.1 & 99.3 & 73.1  &  92.3             \\
    EgoNN\cite{kamorowski2022egonn} & \textbf{100.0}  & 89.3 & 98.4 & \textbf{100.0} & 91.4  & 95.8 \\
		\hline
		BEVPlace~\cite{bevplace}     &{99.7} &{98.1} &{99.3} &\textbf{100.0} &{92.0} &{97.8}  \\
      BEVPlace++ (ours)     &\textbf{100.0} &{\textbf{99.3}} &\textbf{100.0} &\textbf{100.0} & {\textbf{99.1}} & \textbf{99.7}  \\
		\hline
	\end{tabular}
    \vspace{-0.3cm}
	\label{tab: kitti_generalization}
\end{table}
\begin{table}[t]\footnotesize
	\centering
    \renewcommand\arraystretch{1.1}
	\renewcommand\tabcolsep{5pt}
        \captionsetup{aboveskip=0pt, belowskip=0pt}
	\caption{Recall at Top-1 on the rotated KITTI dataset.}% 
	\begin{tabular}{l|cccccc}
		\hline
		Sequence & 00 & 02 & 05 & 06 & 08 & Mean \\ \hline
        M2DP \cite{kim2018scan} & 92.9 & 69.3 & 80.7 & 94.8 & 34.4 & 74.4 \\
        BoW3D     \cite{liu2019lpd}         &19.2&9.1&13.5&13.4 & 1.5 & 11.3 \\
		Logg3D\cite{logg3dnet}  &99.4 & 96.4 & 97.3 & 99.6 & 92.0 &  96.9 \\
		CVTNet     \cite{ma2023cvtnet}     &    98.7 & 87.4 & 93.3 & 98.5 & 85.8 &  92.7\\
     RING++~\cite{xu2023ringplus} &  94.3 &  67.0 &  87.9 &  95.3 &  59.5 &   80.8  \\
		LCDNet  \cite{mickloc3d}       &\textbf{99.7} & \textbf{98.1} & 95.5 & \textbf{100.0} & 94.7 &  97.6 \\
        LCRNet\cite{lcrnet} & 98.4 & 82.3 & \textbf{99.3} & 98.1 & 67.1 & 89.0                \\
        EgoNN\cite{kamorowski2022egonn} & 99.1  & 91.3 & 98.0 & \textbf{100.0} & 86.6 &  95.0 \\
		\hline
		BEVPlace~\cite{bevplace}     &{99.6} &{93.5} &{{98.9}} &{\textbf{100.0}} &{92.0} &{96.8} \\
      BEVPlace++ (ours)     &\textbf{99.7}& 97.1& {98.9}& \textbf{100.0}& \textbf{97.3}&     \textbf{98.6}  \\
		\hline
	\end{tabular}
    \vspace{-0.3cm}
	\label{tab: kitti_rotation}
\end{table}

In this section, we conduct experiments to evaluate the performance of our method in terms of place recognition, loop closing, and global localization. \revisesec{We compare our method with state-of-the-art methods including M2DP \cite{m2dp}, CVTNet \cite{ma2023cvtnet}, Logg3D-Net \cite{logg3dnet}, BoW3d \cite{bow3d}, RING++~\cite{xu2023ringplus}, LCDNet \cite{lcdnet}, LCRNet~\cite{lcrnet}, and EgoNN~\cite{kamorowski2022egonn}. Among them, BoW3d, RING++, LCDNet, LCRNet, and LCRNet can estimate poses, while the other 3 methods can only perform place recognition and loop closure detection.}

\textbf{Implementation details}. For all baseline methods, we reproduce their results using their open-source implementations with default setups\footnote{\parindent=12pt\relax https://github.com/LiHeUA/M2DP,\par https://github.com/BIT-MJY/CVTNet,\par https://github.com/YungeCui/BoW3D,\par https://github.com/csiro-robotics/LoGG3D-Net,\par https://github.com/robot-learning-freiburg/LCDNet,\par https://github.com/nubot-nudt/LCR-Net,\par https://github.com/jac99/Egonn.}. For BEVPlace++, we use ResNet34 as the backbone in REM. The ResNet is cropped to retain the first three layers (up to conv3\_x), resulting in an output channel number $C$ of 128. The number of rotations $N_R$ in REM is empirically set to 8. The point cloud crop range $D$ is set to 40 meters, and the grid size $g$ for BEV image generation is 0.4 meters. Consequently, the BEV image has a size of $200\times 200$. We train BEVPlace++ with the AdamW optimizer for 50 epochs. The learning rate is initially set as 1e-4 and decays by a factor of 0.5 every 10 epochs. The weight decay is set to 1e-3. The method is trained on an RTX 3090 GPU.

\subsection{Place Recognition}

We conduct experiments to fully evaluate the performance of place recognition including the robustness to view changes, generalization ability, and long-term stability. 

\textbf{Performance on KITTI}. We only train the methods on KITTI dataset using the database of sequence ``\textit{00}'', containing 3000 frames. We apply data augmentation to all methods by randomly rotating the point clouds to improve the robustness to view changes. As can be seen in Tab.~\ref{tab: kitti_generalization}, our BEVPlace++ outperforms M2DP, Scan Context, BoW3D, and CVTNet with large margins. \revise{Logg3d-Net, LCDNet, LCRNet, and EgoNN achieve comparable performance to BEVPlace++. However, they perform much worse than BEVPlace++ on the challenging sequence  ``\textit{08}'' which has a large number of challenging reverse loops.}

\textbf{Robustness to view changes.} To test the robustness against rotational changes, we randomly rotate all the query and database point clouds around the z-axis with an angle range of $[0, 2\pi)$ to simulate view changes. As shown in Tab.~\ref{tab: kitti_rotation}, BEVPlace++ maintains the highest recall rates, benefiting from our rotational invariant global descriptor designs. CVTNet, Logg3d-Net, LCDNet, and EgoNN are also robust to rotations to some extent. However, BoW3D's performance significantly degenerated compared to those without rotation changes.

\begin{table*}[t]\footnotesize
	\centering
    \renewcommand\arraystretch{1.1}
	\renewcommand\tabcolsep{3pt}
            \captionsetup{aboveskip=0pt, belowskip=0pt}
	\caption{Generalization Performance of Recall at Top-1 on NCLT and MCD datasets.}% 
	\begin{tabular}{l|cc cc cc| ccccc}
        \hline 
        & \multicolumn{6}{c|}{NCLT} & \multicolumn{5}{c}{MCD\_ntu} \\
		\hline 
                  & \makecell{2012-02-04} & \makecell{2012-03-17} & \makecell{2012-06-15} & \makecell{2012-09-28} & \makecell{2012-11-16} &  \makecell{2013-02-23} & day\_02 & day\_010 & night\_04 & night\_08 & night\_13 \\\hline
	  M2DP~\cite{m2dp}       &  63.2 & 58.0 & 42.4 & 40.6 & 49.3 & 27.9  &  46.7 & 65.5 & 56.1 & 55.7 & 59.8   \\
      BoW3D~\cite{bow3d}  &  14.9 & 10.7 & 6.5 & 5.0 & 5.2 & 7.5   &  0.0 & 0.0 & 0.0 & 0.0 & 0.0            \\
      CVTNet~\cite{ma2023cvtnet} &  89.2 & 88.0 & 81.2 & 74.9 & 77.1 & 80.3   &  80.0 & 84.8 & 82.0 & 83.9 & 85.9       \\
      LoggNet~\cite{logg3dnet} &  69.9 & 19.6 & 11.0 & 8.7 & 10.9 & 25.6  &  6.9 & 12.7 & 8.8 & 10.9 & 10.4          \\
      RING++~\cite{xu2023ringplus}  & 60.0 & 51.3 & 52.1 & 31.2 & 35.5 & 26.9 & 42.8 & 49.5 & 44.1 & 52.2 & 50.2  \\
      LCDNet~\cite{lcdnet} &  60.5 & 54.2 & 44.2 & 34.9 & 31.7 & 10.9 &  45.6 & 53.5 & 50.3 & 46.2 & 48.9    \\
      LCRNet~\cite{lcrnet} &  32.2 & 60.0 & 21.7 & 21.2 & 24.7 & 19.1 &  64.7 & 62.8 & 35.2 & 51.1 & 55.5    \\
      EgoNN~\cite{kamorowski2022egonn} &  44.9 & 43.5 & 41.9 & 34.3 & 31.0 & 28.4 &  55.4 & 73.2 & 36.2 & 66.2 & 64.0    \\
      
      \hline
      BEVPlace~\cite{bevplace} &  93.5 & 92.7 & 87.4 & 87.8 & 88.9 & 86.2   &  79.1 & 87.4 & 80.5 & 85.9 & 83.7         \\
      BEVPlace++ &  \textbf{95.3}  & \textbf{94.2} & \textbf{90.2} & \textbf{88.9} & \textbf{91.3} & \textbf{87.8}  &  \textbf{83.1}  & \textbf{90.2} & \textbf{86.6} & \textbf{88.9} & \textbf{86.4}  \\
	\hline
        \end{tabular}
	\label{tab: nclt_mcd_pr_generalization}
    \vspace{-0.3cm}
\end{table*}

\begin{table*}[!]\footnotesize
	\centering
	%	\vspace{-3mm}
        \captionsetup{aboveskip=0pt, belowskip=0pt}
	\caption{Recall rates on the benchmark dataset.} 
    \renewcommand\arraystretch{1.1}
	\renewcommand\tabcolsep{3pt}
	\begin{tabular}{lcccccccc|cc}
		\hline
		& \multicolumn{2}{c}{\dlmu[2.5cm]{Oxford}}     & \multicolumn{2}{c}{\dlmu[2.5cm]{U.S.}}       & \multicolumn{2}{c}{\dlmu[2.5cm]{R.A.}}       & \multicolumn{2}{c}{\dlmu[2.5cm]{B.D}}        &
		\multicolumn{2}{|c}{Mean}   \\ 
		& AR@1 & AR@1\% & AR@1 & AR@1\% & AR@1 & AR@1\% & AR@1 & AR@1\% & AR@1 & AR@1\%  \\
		\hline
		PointNetVLAD \cite{pointnetvlad}        & 62.8 & 80.3  & 63.2 & 72.6 & 56.1 & 60.3 & 57.2   & 65.3      & 59.8       & 69.6   \\ 
		LPD-Net \cite{liu2019lpd}   & 86.3       & 94.9 & 87.0 & 96.0 & 83.1  & 90.5  & 82.5 & 89.1 & 84.7 & 92.6    \\ 
		NDT-Transformer \cite{ndtformer}      & 93.8       & 97.7          & -          & -      & -       & -       & -       & -     & -       & -  \\ 
		PPT-Net    \cite{pptnet}           & 93.5       & 98.1       & 90.1       & 97.5   & 84.1       & 93.3       & 84.6       & 90.0    & 88.1       & 94.7   \\ 
		SVT-Net    \cite{svtnet}             & 93.7       & 97.8       & 90.1       & 96.5   & 84.3       & 92.7       & 85.5       & 90.7    & 88.4       & 94.4   \\ 
		TransLoc3D \cite{transloc3d}         & 95.0       & 98.5          & -          & -      & -       & -       & -       & -    & -       & -   \\ 
		MinkLoc3Dv2  \cite{mickloc3dv2}            & 96.3       & 98.9       & 90.9       & 96.7   & 86.5       & 93.8       & 86.3       & 91.2    & 90.0       & 95.1   \\ 
		\hline
		BEVPlace~\cite{bevplace}   & \textbf{96.5}   & {99.0}   & {96.9}   & {\textbf{99.7}}   & {92.3}         & {98.7}   & {95.3}         & {99.5}   & {95.3}       & {99.2}  \\
        BEVPlace++ (ours)  & {96.2}   & {\textbf{99.1}}   & {\textbf{97.1}}   & {\textbf{99.7}}   & {\textbf{92.7}}         & {\textbf{98.8}}   & {\textbf{95.6}}         & {\textbf{99.6}}   & {\textbf{95.4}}       & {\textbf{99.3}}  \\
		\hline %
	\end{tabular}
	\label{tab: recall_oxford}
    \vspace{-0.5cm}
\end{table*}

\begin{table*}[!htp]\footnotesize
	\centering
        \renewcommand\arraystretch{1.2}
	\renewcommand\tabcolsep{2.5pt}
        \captionsetup{aboveskip=0pt, belowskip=0pt}
	\caption{Average precision and F1 max score of loop closure on KITTI and NCLT. \revisesec{\textdagger\ 3-DoF methods.}} 
	\begin{tabular}{l|cc cc c|ccc cc | cc  cc c | c cc cc}
		\hline
		Sequence & \multicolumn{5}{c|}{KITTI 00} & \multicolumn{5}{c|}{KITTI 02} & \multicolumn{5}{c|}{KITTI 05} & \multicolumn{5}{c}{KITTI 06}    \\ \hline

        & AP & \makecell{max \\F1} & \makecell{max \\ Recall \\ (\%)} & \makecell{$\hat{e}_t$ \\ (m)} & \makecell{$\hat{e}_r$\\$ (^\circ)$} & AP & \makecell{max \\F1} & \makecell{max \\ Recall \\ (\%)} & \makecell{$\hat{e}_t$ \\ (m)} & \makecell{$\hat{e}_r$\\$ (^\circ)$} & AP & \makecell{max \\F1} & \makecell{max \\ Recall \\ (\%)} & \makecell{$\hat{e}_t$ \\ (m)} & \makecell{$\hat{e}_r$\\$ (^\circ)$} & AP & \makecell{max \\F1} & \makecell{max \\ Recall \\ (\%)} & \makecell{$\hat{e}_t$ \\ (m)} & \makecell{$\hat{e}_r$\\$ (^\circ)$}  \\\hline

        M2DP \cite{kim2018scan} & 0.982 & 0.936 & 86.7 & - & - & 0.884 & 0.844 & 0.0 & - & - & 0.946 & 0.897 & 68.1 & - & - & 0.974 & 0.938 & 76.2  & - & -  \\
		Logg3D\cite{logg3dnet}  & 0.995 & 0.976 & 55.2 & - & - & 0.983 & 0.927 & 82.7 & - & - & 0.995 & 0.975 & 86.2 & - & - & 0.996 & 0.970 & 91.9  & - & -   \\
		CVTNet     \cite{soe}        & 0.994 & 0.965 & 84.8 & - & - & 0.931 & 0.898 & 64.6 & - & - & 0.975 & 0.933 & 96.2 & - & - & 0.996 & 0.981  & 96.2 & - & -\\
        BoW3D     \cite{liu2019lpd}         &0.979& 0.897 & 46.5 & 0.54 & 1.20 & 0.559 & 0.546 & 10.6 & 0.74 & 0.55 & 0.957 & 0.857 & 47.8 & 0.69 & 0.72 & 0.992 & 0.968 & 48.1 & 0.62 & 0.73 \\
        RING++~\cite{xu2023ringplus}\textdagger & 0.946 & 0.921 & 89.1 & 0.41 & 0.52       & 0.921 & 0.909 &  80.2 & 0.55 & 0.81       & 0.952 & 0.941 & 89.5 & 0.44 & 0.43      & 0.982 & 0.968 & 95.6 & 0.27 & 0.25    \\
		LCDNet  \cite{mickloc3d}       & 0.997 & 0.974 & 94.1 & 0.10 & 0.14 & 0.976 & 0.928 & {83.7} & 0.65 & \textbf{0.44} & 0.994 & 0.964 & 93.0 & \textbf{0.12} & 0.17 &  \textbf{0.999} & 0.997 & 99.6  & \textbf{0.11} & 0.17 \\
          LCRNet  \cite{lcrnet}       &  0.997 & 0.975 & 81.7 & 0.12 & 0.23 & 0.956 & 0.923 & {23.0} & 0.91 & {1.44} & 0.991 & 0.959 & 90.3 & {0.19} & 0.25 &  \textbf{0.999} & 0.996 & 90.7  & {0.14} & 0.16 \\
            EgoNN   \cite{kamorowski2022egonn}       & 0.954 & 0.970 & 85.5 & 0.21 & 0.15 & 0.853 & 0.920 & {\textbf{85.3}} & 0.42 & {0.49} & 0.795 & 0.885 & 73.3 & 0.14 & 0.19 &  {0.989} & 0.993 & 96.7  & {0.13} & 0.09 \\
		\hline
      BEVPlace++\textdagger  &\textbf{0.999}& \textbf{0.995}& \textbf{98.4} & \textbf{0.08} & \textbf{0.11} & \textbf{0.977}& \textbf{0.934} & 70.0 & \textbf{0.38} & 0.70 & \textbf{0.994} &\textbf{0.982} & \textbf{96.2} & \textbf{0.12} & \textbf{0.09} & \textbf{0.999} & \textbf{0.999} & \textbf{100.0} & 0.18 & \textbf{0.08}  \\
        \hline\hline
      &  \multicolumn{5}{c|}{KITTI 08} & \multicolumn{5}{c|}{\makecell{NCLT 2012-01-15}} & \multicolumn{5}{c|}{\makecell{NCLT 2012-02-04}} & \multicolumn{5}{c}{\makecell{NCLT 2012-03-17}}   \\\hline
      & AP & \makecell{max \\F1} & \makecell{max \\ Recall \\ (\%)} & \makecell{$\hat{e}_t$ \\ (m)} & \makecell{$\hat{e}_r$\\$ (^\circ)$} & AP & \makecell{max \\F1} & \makecell{max \\ Recall \\ (\%)} & \makecell{$\hat{e}_t$ \\ (m)} & \makecell{$\hat{e}_r$\\$ (^\circ)$} & AP & \makecell{max \\F1} & \makecell{max \\ Recall \\ (\%)} & \makecell{$\hat{e}_t$ \\ (m)} & \makecell{$\hat{e}_r$\\$ (^\circ)$} & AP & \makecell{max \\F1} & \makecell{max \\ Recall \\ (\%)} & \makecell{$\hat{e}_t$ \\ (m)} & \makecell{$\hat{e}_r$\\$ (^\circ)$}  \\\hline
        M2DP \cite{kim2018scan} & 0.081 & 0.162 & 0.0 & - & -   & 0.783 & 0.695 & 4.8 & - & - 
  & 0.700 & 0.620 & 3.7 & - & -
  & 0.654 & 0.621 & 4.0 & - & - \\
        
		Logg3D\cite{logg3dnet}  & 0.958 & 0.929 & 2.7 & - & -   & 0.679 & 0.592 & 1.0 & - & - 
  & 0.575 & 0.517 & 0.6 & - & -
  & 0.570 & 0.530 & 1.4 & - &  \\
		CVTNet     \cite{soe}          & 0.848 & 0.721 & 26.0 & - & -   & 0.947 & 0.876 & 20.5 & - & - 
  & 0.923 & 0.863 & 30.3 & - & -
  & 0.907 & 0.836 & 11.2 & - & -\\
        BoW3D     \cite{liu2019lpd}         & 0.905 & 0.829 & 14.4 & 1.44 & 2.81 & 0.000 & 0.000 & 0.0 & - & - & 0.000 & 0.000 & 0.0 & - & - & 0.000 & 0.000 & 0.0 & - & -  \\
        RING++~\cite{xu2023ringplus}\textdagger & 0.809 & 0.776 & 15.9 & 1.12 & 1.01  &  0.852 & 0.799 & 16.4 & 1.02 & 1.27     &  0.843 & 0.801 & 25.3 & 0.97  &    0.84&  0.836 & 0.808 & 6.0 & 1.29 & 1.35       \\
		LCDNet  \cite{mickloc3d}       & 0.952 & 0.918 & 12.2 & \textbf{0.21} & \textbf{0.47}   & 0.633 & 0.342 & 0.0 & 0.39 & {1.20} 
  & 0.621 & 0.362 & 0.0 & 0.37 & 1.16
  & 0.684 & 0.321 &  0.0 & \textbf{0.37} & 1.26 \\
  LCRNet  \cite{lcrnet}       & 0.680 & 0.674 & 3.0 & 1.41 & 1.08 & 0.823 & 0.736 & {2.0} & 1.87 & {1.56} & 0.828 & 0.763 & 6.4 & {1.29} & 1.37 &  {0.847} & 0.770 & 2.7  & {1.88} & 1.41 \\
  EgoNN   \cite{kamorowski2022egonn}       & 0.539 & 0.689 & 31.1 & 0.38 & 0.69 & 0.154 & 0.285 & {1.4} & 0.82 & {1.70} & 0.091 & 0.216 & 0.3 & {1.54} & 1.93 &  {0.105} & 0.222 & 0.3  & {1.58} & 1.20 \\
		\hline
       BEVPlace++\textdagger  & \textbf{0.999} &\textbf{ 0.984} & \textbf{76.4} & 0.35 & 0.57 & \textbf{0.963} & \textbf{0.912} & \textbf{24.9} & \textbf{0.34} & \textbf{1.09} & \textbf{0.969} & \textbf{0.916} & \textbf{34.5} & \textbf{0.36} & 1.19 & \textbf{0.935} & \textbf{0.859} & \textbf{31.2} & 0.40 & \textbf{1.17} \\\hline\hline
		%		\bottomrule
  &  \multicolumn{5}{c|}{NCLT 2012-06-15} & \multicolumn{5}{c|}{\makecell{NCLT 2012-09-28}} & \multicolumn{5}{c|}{\makecell{NCLT 2012-11-16}} & \multicolumn{5}{c}{\makecell{NCLT 2013-02-23}}   \\\hline
  
  & AP & \makecell{max \\F1} & \makecell{max \\ Recall \\ (\%)} & \makecell{$\hat{e}_t$ \\ (m)} & \makecell{$\hat{e}_r$\\$ (^\circ)$} & AP & \makecell{max \\F1} & \makecell{max \\ Recall \\ (\%)} & \makecell{$\hat{e}_t$ \\ (m)} & \makecell{$\hat{e}_r$\\$ (^\circ)$} & AP & \makecell{max \\F1} & \makecell{max \\ Recall \\ (\%)} & \makecell{$\hat{e}_t$ \\ (m)} & \makecell{$\hat{e}_r$\\$ (^\circ)$} & AP & \makecell{max \\F1} & \makecell{max \\ Recall \\ (\%)} & \makecell{$\hat{e}_t$ \\ (m)} & \makecell{$\hat{e}_r$\\$ (^\circ)$}  \\\hline

  M2DP \cite{kim2018scan} 

  & 0.666 & 0.617 & 1.9 & - & -
  & 0.676 & 0.602 & 4.2 & - & -
  & 0.281 & 0.380 & 0.0 & - & -
  & 0.700 & 0.656 & 1.3 & - & -\\
        
		Logg3D\cite{logg3dnet}  & 0.427 & 0.413 & 0.3 & - & -
  & 0.509 & 0.476 & 1.0 & - & -
  & 0.282 & 0.279 & 0.0 & - & -
  & 0.511 & 0.472 & 0.2 & - & - \\
		CVTNet     \cite{soe}         & 0.937 & 0.869 & 36.8 & - & -
  & 0.920 & 0.840 & 19.7 & - & -
  & 0.784 & 0.719 & 8.1 & - & -
  & 0.897 & 0.823 & 15.4 & - & -\\
        BoW3D     \cite{liu2019lpd}         & 0.024 & 0.102 & 0.0 & - & - & 0.000 & 0.000 & 0.0 & - & - & 0.000 & 0.000 & 0.0 & - & - & 0.000 & 0.000 & 0.0 & - & -  \\
        RING++~\cite{xu2023ringplus}\textdagger & 0.797 & 0.760 & 29.9 & 1.26 & 1.30  &  0.674 & 0.645 & 11.3 & 1.73 & 1.56   &  0.623  & 0.539 & 5.1 & 1.87 & 2.42 & 0.664 & 0.629 & 6.4 & 1.32 & 1.92  \\
		LCDNet  \cite{mickloc3d}       & 0.628 & 0.288 & 0 & 0.50 & 1.30
  & 0.552 & 0.244 & 0.0 & 0.44 & \textbf{1.27}
  & 0.243 & 0.039 & 0.0 & 0.47 & 1.55
  & 0.231 & 0.191 & 0.0 & 0.52 & 1.68\\
  LCRNet  \cite{lcrnet}       & 0.839 & 0.805 & 2.7 & 1.77 & 1.92 & 0.613 & 0.548 & {4.6} & 1.70 & {1.82} & 0.132 & 0.258 & 0.0 & {1.95} & 2.46 &  {0.482} & 0.603 & 0.0  & {1.79} & 2.04 \\
  EgoNN   \cite{kamorowski2022egonn}       & 0.115 & 0.243 & 0.5 & 1.37 & 1.99 & 0.216 & 0.321 & {0.8} & 1.79 & {1.83} & 0.036 & 0.067 & 0.0 & {1.76} & 2.23 &  {0.010} & 0.101 & 0.0  & {1.63} & 2.30 \\
		\hline
       BEVPlace++\textdagger  & \textbf{0.955} & \textbf{0.901} & \textbf{63.4} & \textbf{0.40} & \textbf{1.19} & \textbf{0.957} & \textbf{0.894} & \textbf{45.3} & \textbf{0.40} & 1.61 & \textbf{0.839} & \textbf{0.733} & \textbf{15.8} & \textbf{0.40} & \textbf{1.10} & \textbf{0.959} & \textbf{0.887} & \textbf{45.5} & \textbf{0.44} & \textbf{1.05} \\	
       \hline
	\end{tabular}
	\label{tab: lc_ap_f1}
    \vspace{-0.3cm}
\end{table*}

\textbf{Generalization ability and Long-term performance.} We evaluate the methods on NCLT and MCD datasets using models trained on KITTI. For NCLT, we construct the database with the sequence ``\textit{2012-01-15}'' that covers most areas of the campus. We then perform place recognition using the point clouds of other sequences, including the one collected in 2013 across a year. For MCD, we build the database with the sequence ``\textit{ntu\_day\_01}'' and perform place recognition using other sequences, including the three night sequences. These two datasets are collected using different types of LiDAR scanners compared to those used in KITTI and their point clouds are sparser. Tab.~\ref{tab: nclt_mcd_pr_generalization} shows the recall rates at Top-1 on the two datasets. As can be seen, BEVPlace++ achieves high recalls on NCLT regardless of season changes. The compared methods rather have much lower recall rates. BEVPlace++ consistently outperforms other methods on MCD with day-night changes.

\begin{figure*}[htbp]
	\begin{center} 
		\includegraphics [width=7.1in]{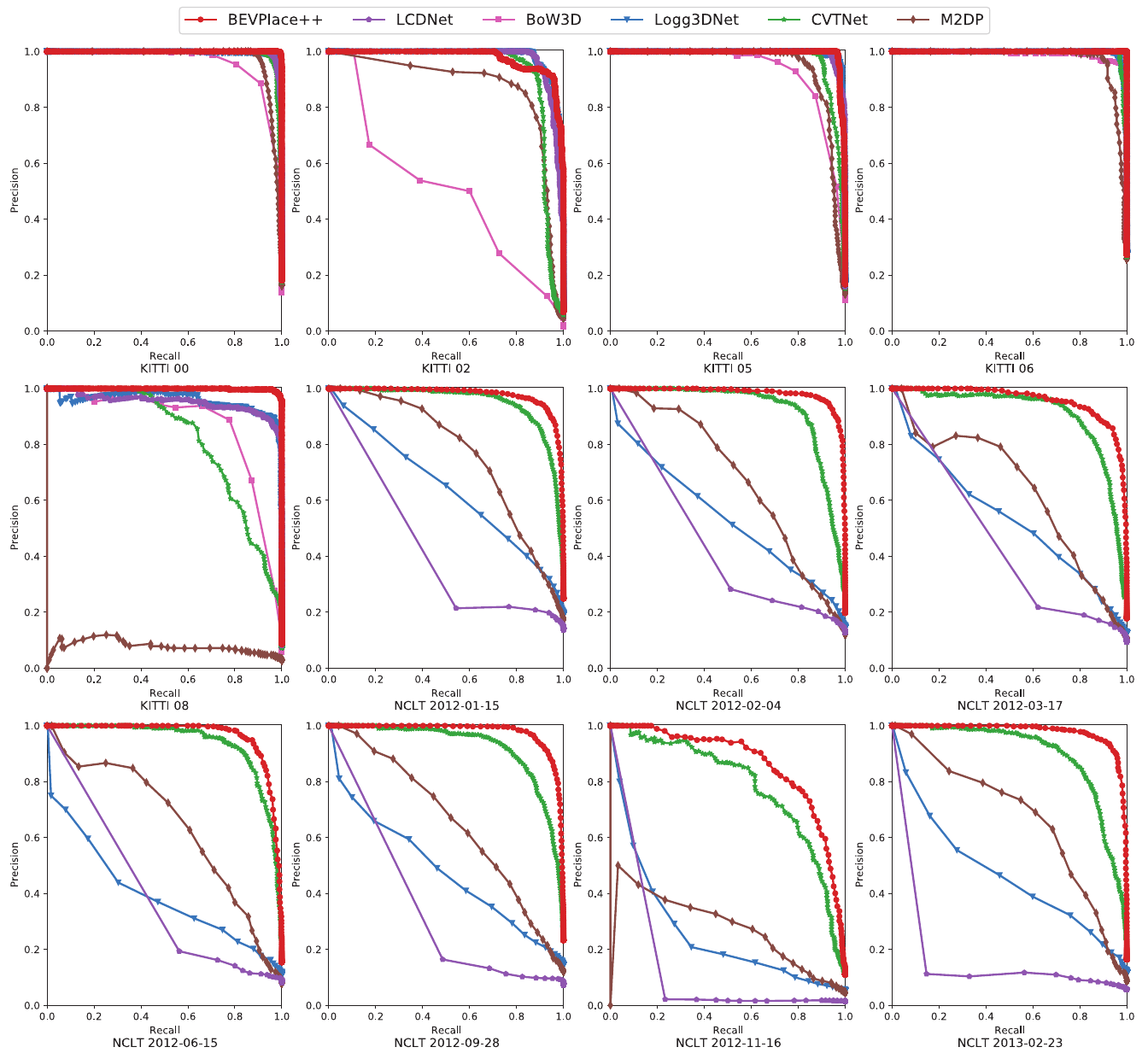}  %[width=0.25\textwidth]
            \captionsetup{aboveskip=0pt, belowskip=0pt}
		\caption{PR curve performance of different methods on KITTI and NCLT datasets.}
		\label{fig: pr_curve}
	\end{center} 
    \vspace{-0.6cm}
\end{figure*}

\textbf{Performance on sparse point maps of RobotCar and Inhouse datasets.} The two datasets provide point clouds downsampled to 4096 points and normalized to range $[-1,1]$. They contain coarse position ground truth. While the compared methods need raw points input or accurate pose supervision, we do not evaluate them on this dataset. Instead, we compare our method with the methods using normalized points, including NDT-Transformer \cite{ndtformer}, PPT-Net \cite{pptnet}, SVT-Net \cite{svtnet}, and TransLoc3D \cite{transloc3d}. For our method, we generate BEV images of size 200$\times$200. Following the previous works~\cite{pointnetvlad,liu2019lpd}, we train our method only with the RobotCar training dataset and test the method on the test set. For all compared methods, we directly use the results from their papers. Tab.~\ref{tab: recall_oxford} shows that our BEVPlace++ outperforms other methods including the transformer-based ones with large margins. In particular, our method generalizes well to U.S, R.A, and B.D subsets, while other methods have relatively large performance degradation.

\begin{table*}[!htp]\footnotesize
	\centering
        \renewcommand\arraystretch{1.2}
	\renewcommand\tabcolsep{1.5pt}
        \captionsetup{aboveskip=0pt, belowskip=0pt}
	\caption{Global localization results on Apollo, Mulran, NCLT, and MCD datasets. \revisesec{\textdagger\ 3-DoF methods.}} 
	\begin{tabular}{l|cc cc |ccc c |  ccc c | cc cc}
		\hline
       & \multicolumn{4}{c|}{\makecell{Apollo}} & \multicolumn{4}{c|}{\makecell{Mulran}} & \multicolumn{4}{c|}{\makecell{NCLT 2012-02-02}} & \multicolumn{4}{c}{\makecell{NCLT 2012-03-17}}   \\\hline
       &\makecell{Recall (\%)} & \makecell{SR (\%)} & \makecell{$\hat{e}_t$ (m)} & \makecell{$\hat{e}_r$ $ (^\circ)$} & \makecell{Recall (\%)} & \makecell{SR (\%)} & \makecell{$\hat{e}_t$ (m)} & \makecell{$\hat{e}_r$ $ (^\circ)$} & \makecell{Recall (\%)} & \makecell{SR (\%)} & \makecell{$\hat{e}_t$ (m)} & \makecell{$\hat{e}_r$ $ (^\circ)$} & \makecell{Recall (\%)} & \makecell{SR (\%)} & \makecell{$\hat{e}_t$ (m)} & \makecell{$\hat{e}_r$ $ (^\circ)$}  \\\hline
      
      BoW3D\cite{bow3d} & 0.0  & 0.0 & - & -   & 3.6    &   1.8   &  0.57  &  0.46  &14.9    &   3.8   &  1.11  &  2.08 &10.7    &   3.0   &  1.02  &  2.44  \\ 
      RING++~\cite{xu2023ringplus}\textdagger  & 83.2 & 76.4  &  0.62 & 0.44  &   79.1 &  65.7 & 0.71 & 0.49  &  60.0 & 56.8 & 1.11 & 1.23    &  51.3 & 42.2 & 1.29 & 1.33  \\
      LCDNet\cite{lcdnet} & 51.1  & 53.4  & 0.56 & 1.01   & 35.8   &     11.2   &  1.09  &  1.29  &60.5 &  58.5   &  0.37  &  1.15 &54.2      &   52.0   &  0.37  &  1.26  \\ 
      LCRNet\cite{lcrnet} & 67.4 & 91.8& \textbf{0.11} & 1.57 & 57.9 & 39.1 & 0.99 & 1.57 & 56.2   &     48.6   &  1.05  &  1.57 & 85.6   &     65.6   &  0.73  &  1.58 \\
      EgoNN\cite{kamorowski2022egonn} & 87.9 & \textbf{93.9} & 0.22& 0.47 &  71.6 & 76.0& \textbf{0.41} & 0.70 & 46.3   &     13.0   &  1.09  &  2.19 & 44.6   &     13.7   &  1.05  &  2.19 \\
      BEVPlace++ & \textbf{88.5} & 81.6 & 0.55 & \textbf{0.29} & \textbf{83.1} & \textbf{80.7} & 0.49 & \textbf{0.40} & \textbf{95.3} & \textbf{95.6}& \textbf{0.32} & \textbf{1.06} & \textbf{94.2} & \textbf{95.1}&\textbf{0.33}&\textbf{1.18} \\\hline\hline
      
      & \multicolumn{4}{c|}{\makecell{NCLT 2012-06-15}} & \multicolumn{4}{c|}{\makecell{NCLT 2012-09-28}} & \multicolumn{4}{c|}{\makecell{NCLT 2012-11-16}} & \multicolumn{4}{c}{\makecell{NCLT 2013-02-23}} \\\hline
       &\makecell{Recall (\%)} & \makecell{SR (\%)} & \makecell{$\hat{e}_t$ (m)} & \makecell{$\hat{e}_r$ $ (^\circ)$} & \makecell{Recall (\%)} & \makecell{SR (\%)} & \makecell{$\hat{e}_t$ (m)} & \makecell{$\hat{e}_r$ $ (^\circ)$} & \makecell{Recall (\%)} & \makecell{SR (\%)} & \makecell{$\hat{e}_t$ (m)} & \makecell{$\hat{e}_r$ $ (^\circ)$} & \makecell{Recall (\%)} & \makecell{SR (\%)} & \makecell{$\hat{e}_t$ (m)} & \makecell{$\hat{e}_r$ $ (^\circ)$}  \\\hline
       BoW3D \cite{bow3d} & 6.5    &   1.1   &  1.23  &  2.62 & 10.7    &   3.0   &  1.02  &  2.44 & 5.2    &   0.3   &  1.27  &  2.36   & 7.5    &   1.1   &  1.05  &  2.14    \\
    RING++~\cite{xu2023ringplus}\textdagger & 52.1 & 40.3 & 1.25 & 1.68  &     31.2 & 30.3  & 1.32 & 1.75   &  35.5 & 28.7 & 1.29 & 1.71  &  26.9 & 19.1 & 1.33 & 1.75  \\
       LCDNet \cite{lcdnet} & 44.2  & 40.0   &  0.49  &  1.28 & 34.9 & 32.2   &  \textbf{0.44}  &  1.27 & 31.7   &   28.8   &  0.47  &  \textbf{1.54} & 10.9 &   6.8   &  0.50  &  1.62 \\
       LCRNet\cite{lcrnet} & 43.2   &     34.8   &  0.52  &  1.59 & 33.9   &     6.9   &  0.91  &  1.59 & 43.2   &     14.2   &  0.93  &  1.58 & 37.1   &     18.2   &  1.01  &  1.54 \\
      EgoNN\cite{kamorowski2022egonn} & 39.2   &     11.7   &  1.11  &  2.13 & 39.5   &     12.0   &  1.11  &  2.16 & 33.5   &     8.3   &  1.12  &  2.35 & 32.0   &     8.3   &  1.05  &  2.18 \\
      BEVPlace++\textdagger & \textbf{90.2} & \textbf{90.9}& \textbf{0.42}&\textbf{1.11} & \textbf{88.9}&\textbf{89.8}&0.46&\textbf{1.23} & \textbf{91.3} &\textbf{90.2}&\textbf{0.44}&1.65  & \textbf{87.8}&\textbf{88.5}&\textbf{0.37}& \textbf{1.05} \\\hline\hline
		%		\bottomrule
  &  \multicolumn{4}{c|}{MCD ntu\_day\_02} & \multicolumn{4}{c|}{\makecell{MCD ntu\_day\_10}} & \multicolumn{4}{c|}{\makecell{MCD ntu\_night\_04}} & \multicolumn{4}{c}{\makecell{MCD ntu\_night\_08}}   \\ \hline
  &\makecell{Recall (\%)} & \makecell{SR (\%)} & \makecell{$\hat{e}_t$ (m)} & \makecell{$\hat{e}_r$ $ (^\circ)$} & \makecell{Recall (\%)} & \makecell{SR (\%)} & \makecell{$\hat{e}_t$ (m)} & \makecell{$\hat{e}_r$ $ (^\circ)$} & \makecell{Recall (\%)} & \makecell{SR (\%)} & \makecell{$\hat{e}_t$ (m)} & \makecell{$\hat{e}_r$ $ (^\circ)$} & \makecell{Recall (\%)} & \makecell{SR (\%)} & \makecell{$\hat{e}_t$ (m)} & \makecell{$\hat{e}_r$ $ (^\circ)$}  \\\hline
  BoW3D \cite{bow3d} & 0.0 & 0.0 & - & - & 0.0 & 0.0 & - & - & 0.0 & 0.0 & - & - & 0.0 & 0.0 & - & - \\
  RING++~\cite{xu2023ringplus}\textdagger & 42.8 & 40.1 & 1.22 & 1.31  &  49.5 & 42.4 & 1.13 & 1.37 & 44.1 & 40.2 & 1.34 & 1.50  &  52.2 & 49.3 & 1.24 & 1.33  \\
  LCDNet \cite{lcdnet} & 45.6   &     29.9   &  1.06  &  \textbf{0.91} & 53.5   &     37.3   &  0.88  &  \textbf{0.90} & 50.3   &     30.3   &  1.01  &  \textbf{0.98} & 46.2   &     33.7   &  0.89  &  \textbf{0.94} \\
  LCRNet\cite{lcrnet} & 81.5   &     76.3   &  \textbf{0.41}  &  1.57 & 75.7   &     70.8   &  \textbf{0.50}  &  1.57 & 50.3   &     45.0   &  \textbf{0.44}  &  1.57 & 63.2   &     58.4   &  \textbf{0.50}  &  1.57 \\
  EgoNN\cite{kamorowski2022egonn} & 60.2   &     18.4   &  1.12  &  2.39 & 59.1   &     27.0   &  1.07  &  2.17 & 30.0   &     10.6   &  0.99  &  2.33 & 60.0   &     27.3   &  1.04  &  2.22 \\
  BEVPlace++\textdagger & \textbf{83.1} & \textbf{77.9}&{0.75}&1.08 & \textbf{90.2} & \textbf{88.8}&{0.61}&1.03  & \textbf{86.6}&\textbf{80.9}&{0.62}& 1.01  & \textbf{88.9}&\textbf{85.7}&{0.64}&1.08 \\
       % \bottomrule 
       \hline
	\end{tabular}
    \vspace{-0.3cm}
	\label{tab: complete_localization}
\end{table*}

\subsection{Loop Closing}
Similar to the setup in place recognition, we test the methods with models trained on sequence ``\textit{00}'' of KITTI. We perform evaluation on every single sequence. For each frame in a sequence, we perform place recognition in the former frames with the nearest 100 frames excluded~\cite{overlapnet,overlaptransformer}. 

Tab.~\ref{tab: lc_ap_f1} presents a quantitative comparison of average precision, F1 max score, max recall at 100\% precision, mean translation errors, and mean rotation errors. BEVPlace++ achieves an average precision of over 90\% on both datasets. While LCDNet and Logg3DNet also show high precision on KITTI, their precision and F1 max scores are significantly lower on NCLT. LCRNet and EgoNN struggle with reverse loops and have relatively lower precision on sequence ``08" of KITTI. Notably, BEVPlace++ achieves a significantly higher maximum recall at 100\% precision on sequence "08" compared to other methods. This high precision is crucial because false positive detections can introduce irreversible errors into downstream tasks such as loop correction and map updating. Ensuring a high recall at high precision means that BEVPlace++ can reliably recognize places without mistakenly identifying incorrect matches, thus maintaining the integrity and accuracy of subsequent processes in autonomous navigation systems. Additionally, BEVPlace++ demonstrates low mean translation and rotation errors on both datasets, specifically below 0.5 meters and 1.5 degrees, respectively. These small errors are significant because they indicate a high level of initial accuracy in the localization process. 
The full PR curves are illustrated in Fig. \ref{fig: pr_curve}. As shown, BEVPlace++ outperforms all baseline methods in terms of PR curve evaluation. Especially on NCLT, BEVPlace++ has a much higher curve.

\subsection{Global Localization}
This experiment evaluates the accuracy of global localization, estimating 3-DoF poses against a pre-built map without knowing the initial pose.
For each query, we retrieve its Top-1 match from the database via place recognition and then compute the global pose using pose estimation. Tab.~\ref{tab: complete_localization} shows the place recognition recall, localization success rates, mean translation error, and mean rotation error. Our BEVPlace++ achieves the highest recall on all sequences and generalizes better on Mulran, NCLT, and MCD compared to BoW3d, LCDNet, LCRNet, and EgoNN, demonstrating its superiority across different sensor configurations and diverse environments. Notably, the localization success rate of BEVPlace++ on NCLT can be higher than the retrieval recall rate, when BEVPlace++ successfully estimates the pose of a query even when the distance between the query and the Top-1 match is larger than 5 meters. It should be noted that BEVPlace++ does not use pose supervision, making it much easier for deployments than methods like LCDNet, LCRNet, and EgoNN.

\subsection{Runtime Analysis}
We compare the runtime of the methods on a desktop equipped with an RTX 3090 GPU and an Intel quad-core 3.40 GHz i5-7500 CPU. Tab.~\ref{tab: runtime} shows the running time of each stage on the KITTI dataset. BEVPlace++ comprises simple residual and NetVLAD blocks, achieving an average frequency of over 40 Hz for place recognition. For global localization, it takes 41.6 ms for pairwise feature extraction, 0.24 ms for place recognition, and 12.7 ms for pose estimation, achieving a frequency of over 20 Hz. As the frequency of LiDAR scans is usually set to 10 Hz, BEVPlace++ can operate in real time.

\begin{table}[t]\footnotesize
	\centering
    \renewcommand\arraystretch{1.1}
	\renewcommand\tabcolsep{6pt}
        
        \captionsetup{aboveskip=0pt, belowskip=0pt}
	\caption{Inference Time on The KITTI dataset.} 
	\begin{tabular}{l|cccc}
		\hline
		  & \thead{Feature \\ Extract. (ms)} & \thead{Place \\ Recognition (ms) } & \thead{Pose \\ Estim. (ms) }  \\ \hline
        M2DP \cite{kim2018scan} & 395.6 & 0.02 & -    \\
		Logg3D\cite{logg3dnet}  &47.3&0.06&-   \\
		CVTNet     \cite{soe}         &9.24&0.01&- \\
        BoW3D     \cite{liu2019lpd}         &80.4 & 10.5& 40.0  \\
        
		LCDNet  \cite{mickloc3d}       &201.5&0.02& 297.0   \\
        LCRNet \cite{lcrnet}  & 289.3  & 0.02 &  342.9   \\
        EgoNN \cite{kamorowski2022egonn}  & 24.6 & 0.06 & 5.7 \\
		\hline
		BEVPlace++     &{20.8} &{0.24} &{12.7} \\
		%		\bottomrule
		\hline
	\end{tabular}
    \vspace{-0.5cm}
	\label{tab: runtime}
\end{table}

In addition to achieving real-time performance, BEVPlace++ is also highly lightweight. The BEVPlace++ model has a size of just 17 MB, which is significantly smaller compared to LCDNet (138 MB), LCRNet (96 MB), and EgoNN (19 MB). Furthermore, the BEV image format averages approximately 20.4 KB per frame, substantially reducing memory consumption. In contrast, LCDNet, LCRNet, and EgoNN store raw point cloud data, requiring around 4.0 MB per frame on the KITTI dataset.

\begin{figure}[t]
	\begin{center} 
		\includegraphics [width=3.5in]{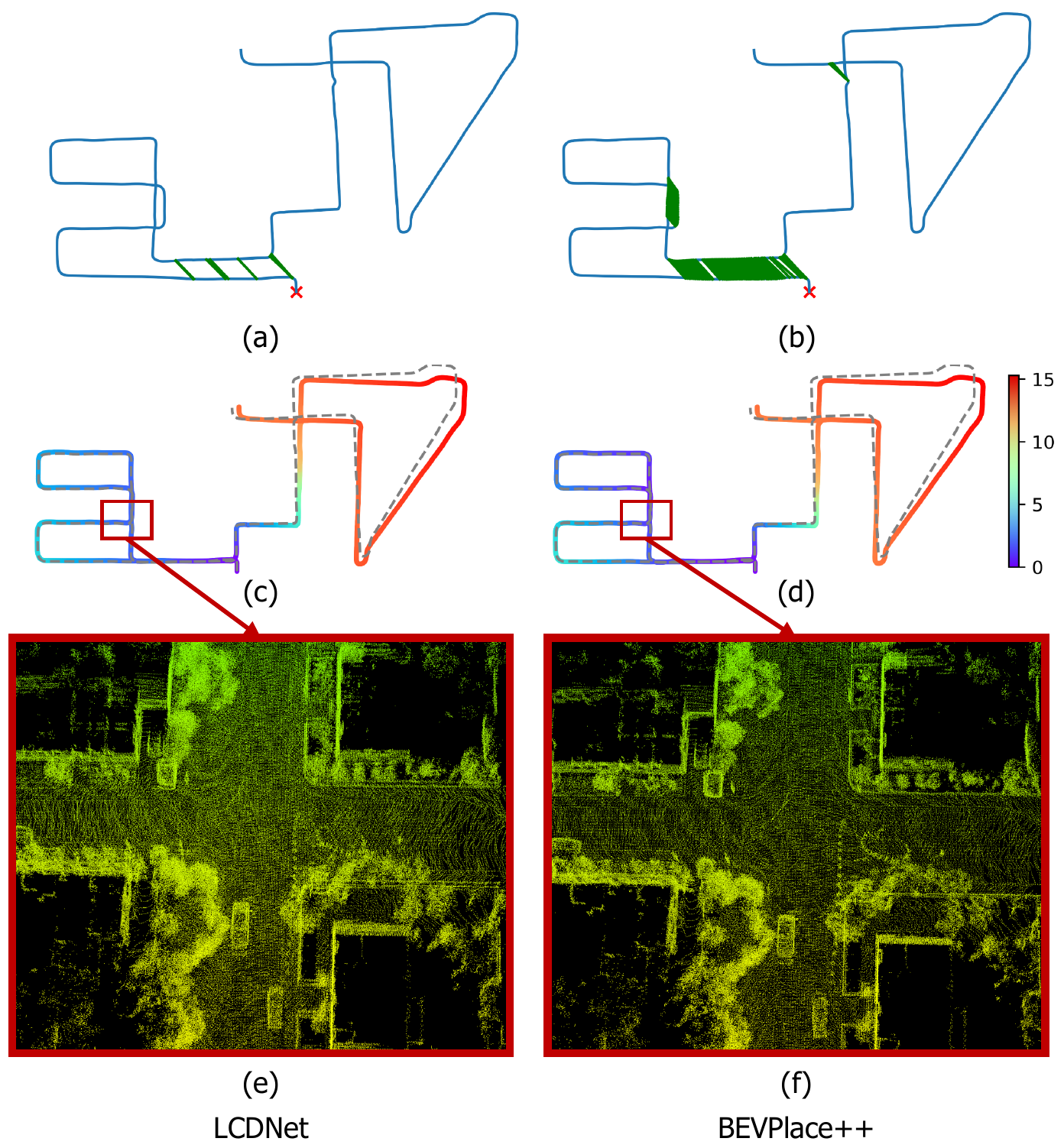}  %[width=0.25\textwidth]
            \captionsetup{aboveskip=0pt, belowskip=0pt}
		\caption{ Performance of A-LOAM with the LCDNet (left) compared to BEVPlace++ (right) on sequence 08 of the KITTI dataset. (a), (b) are detected loops (green lines) under 100\% precision. \textcolor{red}{x} indicates the start point of the trajectory. (c), (d) show the absolute translation errors of LCDNet and BEVPlace++, respectively. (e), (f) are the optimized point cloud map by two methods at the same crossroads.}
		\label{fig: application}
	\end{center} 
    \vspace{-0.7cm}
\end{figure}

\subsection{Application}\label{application}
We evaluate the loop closing performance of BEVPlace++ and LCDNet by integrating these methods into a LiDAR SLAM system, i.e., A-LOAM. We conduct our evaluation on sequence ``08'' of the KITTI dataset, which contains challenging reverse loops. \revise{In the experiments, we generate BEV images from the incoming LiDAR scans. Local and global descriptors are computed by BEVPlace++, which are then associated with their corresponding scan poses from the SLAM system. The BEV database is constructed by combining local descriptors, global descriptors, and poses. Once the robot traverses over 100 meters~\cite{overlapnet}, loop detection is initiated. For every new scan, we compare its global descriptor with descriptors in the existing BEV database. A scan is identified as a loop candidate if the minimum descriptor distance is below a predefined empirical threshold (set to 0.5). Once detecting a valid loop, relative poses between the current scan and the candidate are computed using the local descriptors.} \revisesec{Under the plane-motion assumption, the roll, pitch, and vertical (z-axis) offsets between the query and the matched loop scan are assumed to be negligible during loop detection. Thus, we use the estimated 3-DoF relative pose (x, y, yaw) as the initial guess for the ICP alignment between the two point clouds, with the remaining three DoF (z, roll, pitch) set to zero. The ICP refinement produces a full 6-DoF relative transformation.} \revise{When a loop closure is confirmed, a loop constraint is added to the pose graph optimization framework implemented with Ceres~\cite{agarwal2012ceres}. The pose graph optimization propagates corrections to prior SLAM poses, improving the consistency of the SLAM results.}

In loop closing, it is crucial to detect more true loops without false positives. Therefore, we adopt the maximum recall at 100\% precision as a criterion and only use true positive loops under this condition. Fig.\ref{fig: application} (a) and (b) show the detected loops of LCDNet and BEVPlace++, respectively. As seen, BEVPlace++ detects more loops than LCDNet. This higher detection rate indicates that BEVPlace++ is more effective at identifying true loop closures, which is essential for maintaining the integrity and accuracy of the overall localization and mapping system. Fig.\ref{fig: application} (c) and (d) illustrate the absolute translation errors of the SLAM trajectories after pose graph optimizations. BEVPlace++ achieves better accuracy. Fig.\ref{fig: application} (e) and (f) show the optimized point cloud maps. As can be seen, the road in Fig.\ref{fig: application} (f) is clearer, and the walls are sharper, validating that the map optimized with loops from BEVPlace++ is more accurate.

\begin{figure}[t]
	\begin{center} 
		\includegraphics [width=\columnwidth]{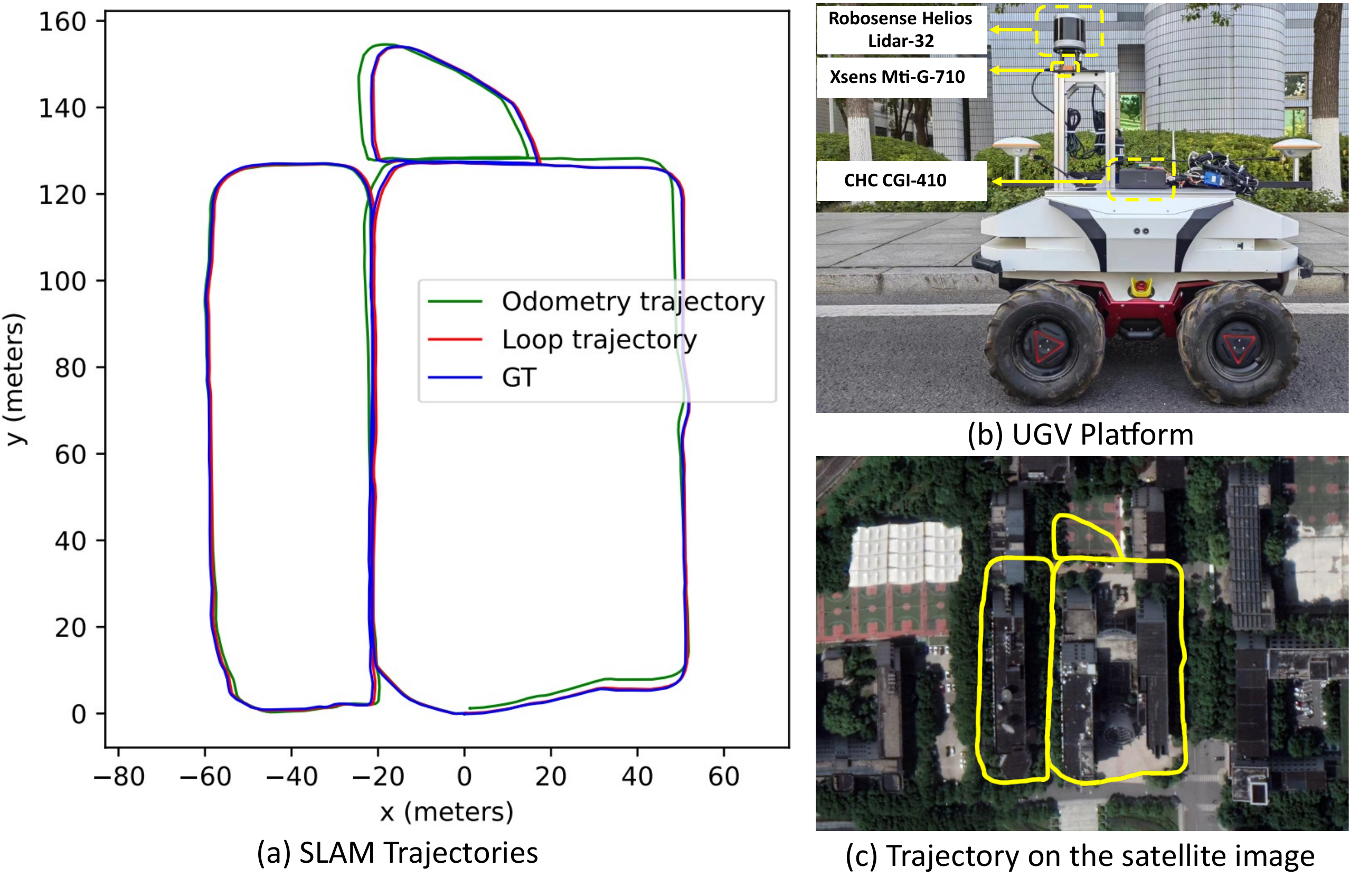}  %[width=0.25\textwidth]
            \captionsetup{aboveskip=0pt, belowskip=0pt}
		\caption{\revise{Comparison of the Odometry trajectory of A-LOAM and the loop trajectory after loop correction using BEVPlace++ on our UGV platform.}}
		\label{fig: nudt_traj}
	\end{center}  
    \vspace{-0.3cm}
\end{figure}

\revise{We conducted additional experiments to evaluate the loop closure performance of BEVPlace++ on a real UGV platform. As shown in Fig.~\ref{fig: nudt_traj}, our UGV was equipped with a RoboSense-32 LiDAR sensor, and its ground truth poses were obtained using a high-accuracy RTK-GPS system. A SLAM method, i.e., A-LOAM integrated with BEVPlace++ was deployed on the robot to construct a globally consistent map. During this process, we evaluated both the loop closure effectiveness and runtime performance of BEVPlace++.}

\revise{As shown in Fig.~\ref{fig: nudt_traj}, the odometry trajectory of A-LOAM has a significant offset from the ground truth. However, after integrating our BEVPlace++ into the SLAM framework and performing loop correction, the trajectory aligns much more closely with the ground truth, indicating improved localization accuracy. After loop closing, the absolute trajectory error was reduced from 0.52 m to 0.29 m, demonstrating that the cumulative drift is significantly reduced. For a comparative analysis, we compared BEVPlace++ with other methods on the same data using metrics including average precision (AP), maximum F1 score, maximum recall at 100\% precision, mean translation error ($e_t$), and mean rotation error ($e_r$). As presented in Tab.~\ref{tab: self_collected}, BEVPlace++ consistently outperformed competing approaches across all metrics, demonstrating its superior generalization capability and effectiveness in real-world scenarios.}

\revise{The proposed method was implemented on an NVIDIA AGX ORIN embedded system. The descriptor extraction and loop candidate search require about 94.9 ms and 1.2 ms, respectively, while RANSAC pose estimation takes 15.1 ms. These results demonstrate that our method operates in real-time on the embedded system, making it both efficient and practical for real-world applications. Furthermore, by implementing descriptor extraction, candidate retrieval, and pose estimation processes in parallel using ROS, the pipeline achieves a real-time performance frequency of up to 10 Hz. A more detailed visualization video of our experiment is included in the multimedia material\footnote{https://youtu.be/4i-ZcqwmMJo?si=TkAp-ymskXTMMjCe}.}

\begin{table}[t]\footnotesize
	\centering
    
    \renewcommand\arraystretch{1.3}
	\renewcommand\tabcolsep{4.5pt}
	\caption{Loop closing performance on the UGV platform.}% * denotes that pose estimation is integrated into the method.} 
	% \vspace{-2mm}
	\begin{tabular}{l|c|c|c|c|c}
        \hline 
        % & \multicolumn{4}{c}  \\
                  & AP & max F1 & \makecell{max recall \\ @ 100\% precision} & $e_t$ (m) & $e_r$ ($\circ$)\\\hline
      \revisesec{LCDNet~\cite{lcdnet}}      &  0.949   &  0.867 & 17.5 & 0.70 & 1.59                                 \\      
      \revisesec{LCRNet~\cite{lcrnet}}  & 0.973 & 0.921 & 37.8 & 0.59 & 1.42     \\
      \revisesec{EgoNN~\cite{kamorowski2022egonn}} &  0.922 & 0.826 & 57.4 & 0.54 & 1.31       \\
      \revisesec{BEVPlace++} &  \textbf{0.999} &  \textbf{0.982} & \textbf{94.9} &  \textbf{0.41} & \textbf{0.98}  \\
	\hline
        \end{tabular}
        \vspace{-3mm}
	\label{tab: self_collected}
\end{table}

\subsection{Understanding BEVPlace++}
\textbf{The backbone choice of REM.} We conduct experiments to explore the performance of BEVPlace++ with different designs in the feature encoder. We replace the local feature encoder REM with a ResNet34~\cite{resnet} and test the method on the rotated KITTI dataset to validate the significance of the rotation equivariance design. Additionally, we use different backbone CNNs in REM to study the robustness of BEVPlace++ to various backbones. Tab.~\ref{table: backbone_ablation} presents the place recognition and pose estimation results of BEVPlace++ on KITTI. From Tab.~\ref{table: backbone_ablation}, we can conclude two observations:
\begin{itemize}
    \item The REM is crucial for achieving robust pose estimation. As shown in Tab.~\ref{table: backbone_ablation}, BEVPlace++ attains higher recall rates for retrieval when using REM encoders and achieves moderate recall rates when using the ResNet encoder. However, the success rates of pose estimation drop significantly when using ResNet34 alone without our REM. This is expected, as the CNN feature map changes considerably with the orientation of the BEV image, making feature matching more challenging. In contrast, the rotation equivariance design of REM aids in pose estimation under large view changes, as the REM feature map is robust to rotational transformations.
    \item BEVPlace++ is robust to the choice of CNN backbone in REM. It achieves the best place recognition recalls and pose estimation success rates when using ResNet34 as the backbone in REM. However, the performance differences when using VGG16~\cite{vgg}, MobileNet~\cite{mobilenetv2}, and EfficientNet~\cite{tan2019efficientnet} as CNN backbones in REM are not significant.
\end{itemize}

\revise{\textbf{Different rotation equivariant encoders for BEVPlace++}. We also conduct an additional ablation study on the KITTI, NCLT, MCD, and Mulran datasets for comparing different rotation equivariant feature encoders in BEVPlace++, including group convolution~\cite{gift} used in BEVPlace, a commonly used rotation equivariant library E2CNN~\cite{e2cnn}, and our proposed REM. As shown in Tab.~\ref{tab: nclt_global_generalization}, BEVPlace++ achieves the highest recall, success rate of global localization (SR), absolute translation error ($e_t$), and absolute rotation error ($e_r$) when using REM, highlighting the effectiveness of our designed REM in preserving feature distinctiveness.}

\begin{table}[!htbp]
    \footnotesize
    \renewcommand\arraystretch{1.1}
    \renewcommand\tabcolsep{8pt}
    \captionsetup{aboveskip=0pt, belowskip=0pt}
    \begin{center}
	\caption{Place recognition and pose estimation performance of BEVPlace++ using different feature encoders}
	\label{table: backbone_ablation}
	\begin{tabular}{c| c c ccc c c c}
            \hline
              &  \multicolumn{4}{c}{KITTI 00}\\\hline
              & Recall(\%) & SR(\%) & $\bar{e}_t$ (m)  & $\bar{e}_r$ ($^\circ$)  \\\hline
            {ResNet34} & 99.4 & 64.7 & 0.95 & 1.13 \\\hline
            {REM(VGG16)}  &  {99.6} & {100.0} &  {\textbf{0.15}} &  {\textbf{0.12}}  \\ 
            {REM(MoblieNet)}  &   {99.4} &  {99.7} &  {0.29} &  {0.29}  \\ 
            {REM(EfficientNet)}  &   {99.4} &  {99.8} &  {0.32} &  {0.18}  \\ \hline
            {REM (ResNet34)}  &   {\textbf{99.7}} &  {\textbf{100.0}} &  {0.16} &  {0.17}  \\ 

            \hline\hline
            &  \multicolumn{4}{c}{KITTI 08}\\\hline
              & Recall(\%) & SR(\%) & $\bar{e}_t$ (m)  & $\bar{e}_r$ ($^\circ$)  \\\hline
            {ResNet34} & 92.5 & 60.2 & 0.77 & 1.01  \\\hline
            {REM(VGG16)}  &   {96.6} &  {97.0} &  {0.58} &  {0.61}  \\ 
            {REM(MoblieNet)}  &   {94.5} &  {96.8} &  {0.62} &  {0.72}  \\ 
            {REM(EfficientNet)}  &   {94.9} &  {95.7} &  {0.59} &  {0.69}  \\ \hline
            {REM (ResNet34)}  &  {\textbf{97.3}} &  {\textbf{98.5}} &  {\textbf{0.54}} &  {\textbf{0.57}}  \\ 
            
            \hline
		\end{tabular}
	\end{center}
    \vspace{-0.3cm}
\end{table}

\begin{table*}[!h]\footnotesize
	\centering
    
    \renewcommand\arraystretch{1.1}
	\renewcommand\tabcolsep{6pt}
        \captionsetup{aboveskip=0pt, belowskip=0pt}
	\caption{Place recognition and pose estimation performance of BEVPlace++ using different rotation equivariant feature encoders.}% 
	\begin{tabular}{l|cc cc |cc cc |cccc|cccc}
        \hline 
        & \multicolumn{4}{c|}{KITTI 08} & \multicolumn{4}{c|}{NCLT 2012-02-15} & \multicolumn{4}{c|}{MCD ntu\_day\_02} & \multicolumn{4}{c}{Mulran}  \\
		\hline 
                  & \makecell{Recall \\ (\%)} & \makecell{SR \\(\%)} & \makecell{$e_t$\\ (m)} & \makecell{$e_r$\\ ($^\circ$)} & \makecell{Recall \\ (\%)} & \makecell{SR \\(\%)} & \makecell{$e_t$\\ (m)} & \makecell{$e_r$\\ ($^\circ$)}  & \makecell{Recall \\ (\%)} & \makecell{SR \\(\%)} & \makecell{$e_t$\\ (m)} & \makecell{$e_r$\\ ($^\circ$)}  & \makecell{Recall \\ (\%)} & \makecell{SR \\(\%)} & \makecell{$e_t$\\ (m)} & \makecell{$e_r$\\ ($^\circ$)}\\\hline
      E2CNN~\cite{e2cnn}      &  91.4   &  94.7 & 0.67 & 0.65 &  67.5   &  71.4 & 0.56 & 1.28  &  56.9   &  55.4 & 0.90 & 1.53    &  60.1   &  61.3 & 0.75 & 0.66                                  \\      
      Group Conv~\cite{gift} &  92.0 & 95.7 & 0.58 & 0.62  &  93.5 & 91.3 & 0.54 & 1.21    &  79.1 & 75.2 & 0.83 & 1.30  &  80.9 & 77.8 & 0.63 & 0.58        \\
      REM &  \textbf{97.3} &  \textbf{98.5} &  \textbf{0.54} & \textbf{0.57} &  \textbf{95.3} &  \textbf{95.6} &  \textbf{0.32} & \textbf{1.06} &  \textbf{83.1} &  \textbf{77.9} &  \textbf{0.75} & \textbf{1.08} &  \textbf{83.1} &  \textbf{80.7} &  \textbf{0.49} & \textbf{0.40}   \\

	\hline
        \end{tabular}
	\label{tab: nclt_global_generalization}
    \vspace{-0.3cm}
\end{table*}

\textbf{Parameter sensitivity.} There are two main parameters in the BEVPlace++ network: the number of rotations $N_R$ in the REM module and the number of clusters $K$ in NetVLAD. We design two independent experiments on KITTI, with each experiment varying only one parameter, to discover the influence of these parameters. As shown in Tab.~\ref{table: parameter_of_ren}, the recall rate increases with the number of orientation intervals $N_R$ and tends to saturate when $N_R\geq 8$. This is reasonable since more accurate rotation equivariance for local features is achieved with larger $N_R$. Considering both computational complexity and localization recall, we set $N_R=8$. The recall rate increases with the number of NetVLAD clusters $K$, but does not show significant improvement when $K\geq 64$. Therefore, we set $K=64$.

\begin{table}[t]
	\renewcommand\arraystretch{1.1}
	\renewcommand\tabcolsep{7.5pt}
        \captionsetup{aboveskip=0pt, belowskip=0pt}
	\begin{center}
		\caption{The recall rates of global localization under different parameter settings of REIN}
		\label{table: parameter_of_ren}
		\begin{tabular}{l|ccccc|c}
			\hline
			\multicolumn{7}{c}{The results of REM parameter $N_R$, fix $K=64$ } \\ \hline
			Seq.                 &  00   & 02   & 05   & 06   & 08  & mean \\ \hline
            $N_R=2$              &  99.71& 88.39& 98.65& 99.63& 92.54&     95.78\\ 
            $N_R=4$              &  99.56& 93.87& 99.10& 100.0& 95.22&     97.55\\ 
            $N_R=6$              &  99.85& 94.84& 99.33& 99.63& 97.61&     98.25\\ 
            $N_R=8$              &  99.71& 97.10& 98.88& 100.0& 97.31&     98.60\\ 
            $N_R=10$              &  99.85& 98.71& 99.10& 100.0& 96.52&     98.82\\ \hline \hline
            \multicolumn{7}{c}{The results of the cluster parameter $K$, fix $N_R=8$ } \\ \hline
			Seq.                 &  00   & 02   & 05   & 06   & 08  & mean \\ \hline
            $K=2$&  99.71& 78.71& 96.41& 99.26& 84.48&     91.71\\ 
            $K=16$&  100.0& 94.19& 98.43& 100.0& 97.31&     97.99\\ 
            $K=32$&  100.0& 94.84& 99.33& 100.0& 95.82&     98.00\\ 
            $K=64$              &  99.71& 97.10& 98.88& 100.0& 97.31&     98.60\\ 
            $K=80$              &  99.85& 97.21& 98.66& 99.63& 98.21&     98.71\\ \hline 
		\end{tabular}
	\end{center}
    \vspace{-0.3cm}
\end{table}

\begin{table}[!t]
	\renewcommand\arraystretch{1.1}
	\renewcommand\tabcolsep{5pt}
        \captionsetup{aboveskip=0pt, belowskip=0pt}
	\begin{center}
		\caption{Global localization under different settings of BEV grid}
		\label{tab: parameter_grid}
		\begin{tabular}{c| c c c| c c c}
            \hline
			Seq. & \multicolumn{3}{c|}{00} & \multicolumn{3}{c}{08} \\ \hline
            & SR(\%)& $\bar{e}_t$ (m) & $\bar{e}_r$ ($^\circ$)   & SR(\%) & $\bar{e}_t$ (m) & $\bar{e}_r$ ($^\circ$) \\
			\hline
            $g=0.1$              &  100.0& 0.09 & 0.11 & 99.1 & 0.38 &     0.47\\ 
            $g=0.2$              &  100.0& 0.08 & 0.10 & 98.5 & 0.38 &     0.33 \\ 
            $g=0.4$              &  100.0 & 0.11 & 0.07 & 98.5 & 0.34 & 0.57 \\ 
            $g=0.6$              &  99.0& 0.31 & 0.17& 96.6& 0.42 & 0.60 \\
            $g=0.8$              &  98.7& 0.44 & 0.25& 95.5& 0.63 & 0.87 \\\hline
		\end{tabular}
	\end{center}
    \vspace{-0.5cm}
\end{table}

The grid size $g$ is the key parameter to BEV image generation. We conduct experiments on sequences ``00'' and ``08'' to evaluate the influence of $g$ on global localization. As shown in Tab.~\ref{tab: parameter_grid}, the success rate decreases and the mean translation and rotation errors tend to increase as the grid size gets large. This is intuitive since the size of the BEV image will get small and the image contents will be highly compressed. We set $g\geq0.4$ in our experiments by trading between accuracy and computation complexity.

\begin{figure*}[t]
	\begin{center} 
		\includegraphics [width=5.8in]{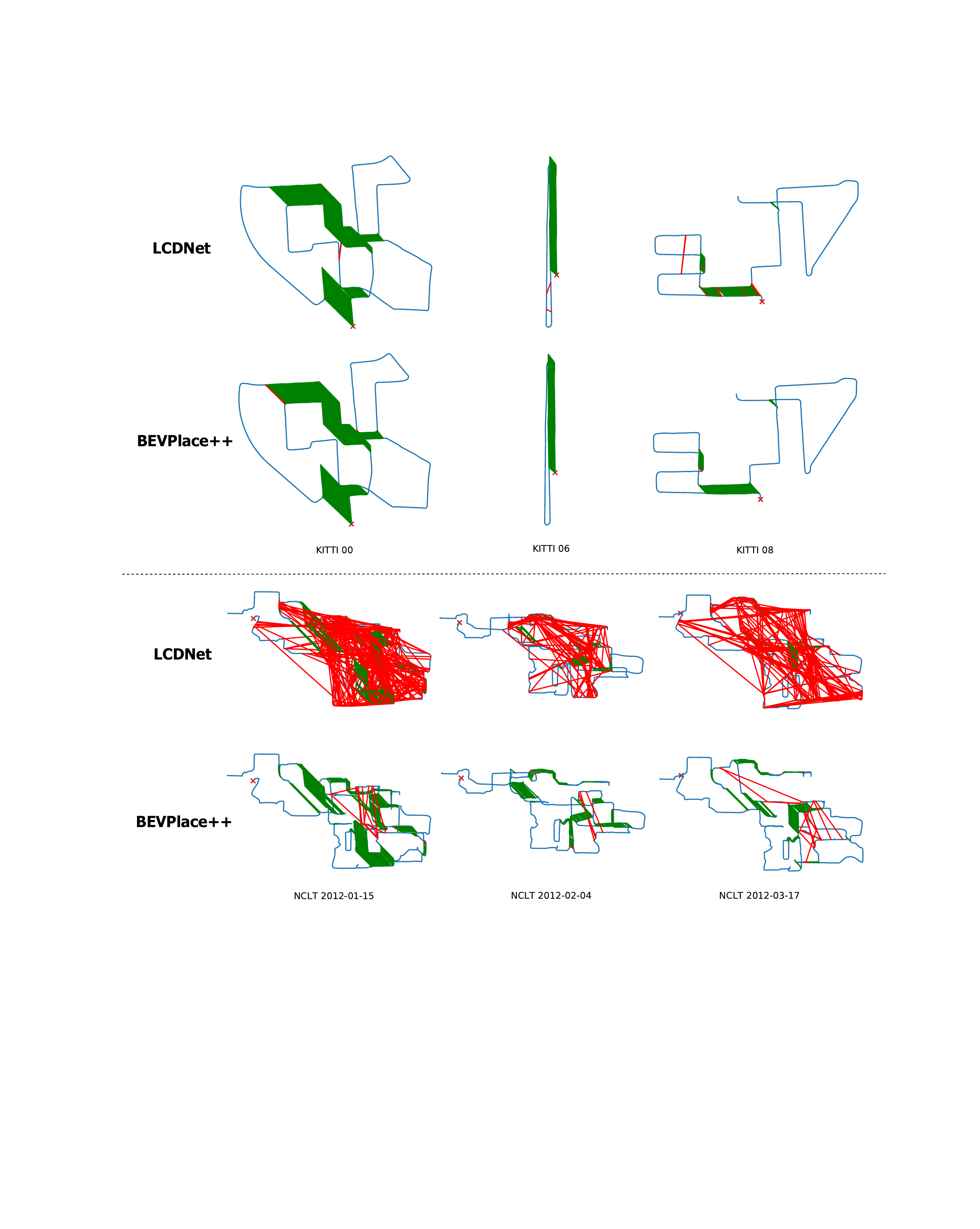}  %[width=0.25\textwidth]
            \captionsetup{aboveskip=0pt, belowskip=0pt}
		\caption{Comparison of detected loops of LCDNet and BEVPlace++ on KITTI and NCLT. The blue line represents the ground truth pose trajectory of the sequence. The red \textcolor{red}{x} marks the starting point of the trajectory. Green lines indicate true positive loop matches, connecting the positions of the query and the matched frames, while red lines represent false positives.}
		\label{fig: loop_detections}
	\end{center}
    \vspace{-0.6cm} 
\end{figure*}

\textbf{More qualitative results.} We present the detected loops of BEVPlace++ and LCDNet on the evaluation sequences in Fig.~\ref{fig: loop_detections}. In these figures, green lines indicate true positives and red lines indicate false positives. On KITTI, BEVPlace++ detects many correct loops. Notably, in the challenging sequence "08", BEVPlace++ successfully detects correct loops in reverse or cross routes. On the contrary, LCDNet detects more false positive loops. On the NCLT dataset, BEVPlace++ detects more false positive loops than on KITTI. The reasons are twofold. First, the point clouds in NCLT are more sparse due to the use of a sparse LiDAR scanner. Second, NCLT contains many challenging areas, such as long corridors and large open spaces, where BEV images lack significant texture information, reducing the distinctiveness of BEV features. Nevertheless, BEVPlace++ generalizes on NCLT much better than LCDNet.

\begin{figure*}[htbp]
	\begin{center} 
		\includegraphics [width=6.0in]{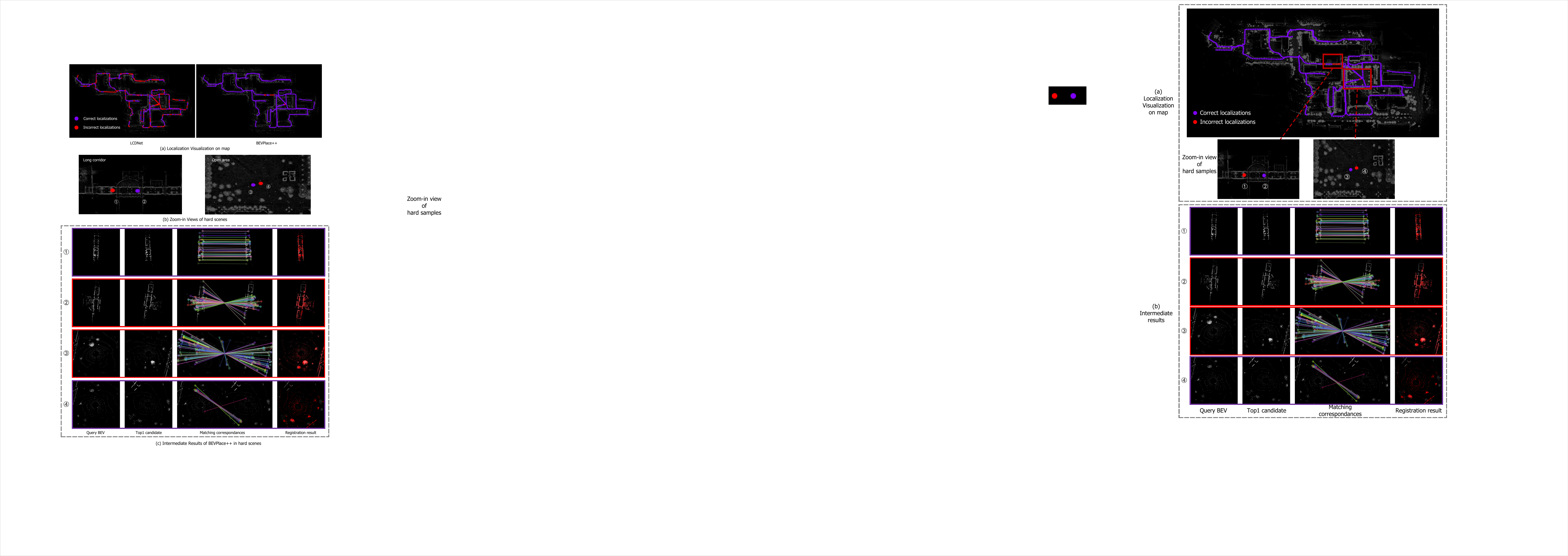}  %[width=0.25\textwidth]
            \captionsetup{aboveskip=0pt, belowskip=0pt}
		\caption{Visulization of the localization results of sequence \textit{2012-02-25} of NCLT. (a) shows the correct (colored purple) and incorrect localizations (colored red) of LCDNet and BEVPlace++. (b) Zoom-in views of hard scenes for localization. (c) The intermediate localization results of BEVPlace++, including the query BEV, the top-1 candidate of place recognition, the feature matching correspondence, and the registration results. The query BEV is colored red for better visualization.}
		\label{fig: localization_hard_samples}
	\end{center} 
    \vspace{-0.6cm}
\end{figure*}

Fig. \ref{fig: localization_hard_samples} (a) visualizes the results of localizing the point clouds from sequence \textit{2012-02-25} on the global map of NCLT using BEVPlace++ and LCDNet. It demonstrates that BEVPlace++ can successfully perform global localization in more areas than LCDNet. The failed localizations (colored red) of BEVPlace++ primarily occur in challenging scenes such as long corridors and open areas with few measurements. We provide a zoomed-in view of these challenging scenes in  Fig. \ref{fig: localization_hard_samples} (b) to better illustrate the difficulties in localization. Fig. \ref{fig: localization_hard_samples} (c) further shows intermediate results of BEVPlace++ in these hard scenes, including the query BEV, the top-1 candidate for place recognition, feature matching correspondence, and warp results. In \textcircled{1} and \textcircled{4} of Fig. \ref{fig: localization_hard_samples} (c), BEVPlace++ retrieves false top-1 candidates for the query. In both cases, the query BEV images lack sufficient structural information, hindering BEVPlace++ from extracting global descriptors with enough distinctiveness. In contrast, \textcircled{2} and \textcircled{3} are examples where BEVPlace++ successfully localizes the queries despite significant view changes.

\section{Conclusion}
 In this paper, we introduce BEVPlace++, a novel global localization method. BEVPlace++ adopts a two-stage method paradigm, sequentially performing place recognition and pose estimation. Utilizing BEV images, BEVPlace++ employs the rotation-equivariant network (REM) to extract robust local features. It generates rotation-invariant global descriptors with NetVLAD pooling. As a global localization method, BEVPlace++ can perform multiple localization tasks, including place recognition, loop closing, and global localization. A key insight of BEVPlace++ is that CNNs inherently extract distinctive features from BEV images, as demonstrated through statistical analysis under translation movements. The proposed REM enhances this distinctiveness under rotational transformations. Leveraging these characteristics, BEVPlace++ enables pose estimation for point clouds without explicit pose supervision and adapts well to diverse LiDAR scanners and unknown environments. We conducted experiments across seven public datasets, showcasing BEVPlace++'s state-of-the-art performance. Additionally, we integrated BEVPlace++ as a loop closing module in a SLAM system, verifying its effectiveness in handling loop closing tasks in a real UGV. BEVPlace++ has been open-sourced to contribute to the robotics community. We hope BEVPlace++ will become a promising new paradigm for LiDAR global localization.

% \newpage
\small{
\bibliography{ref}
\bibliographystyle{ieeetr}
}
\vspace{-0.7cm}
\end{document}